\documentclass[lettersize,journal]{IEEEtran}
\usepackage{amsmath,amsfonts}
\usepackage{algorithmic}
\usepackage{algorithm}
\usepackage{array}
\usepackage[caption=false,font=normalsize,labelfont=sf,textfont=sf]{subfig}
\usepackage{textcomp}
\usepackage{stfloats}
\usepackage{url}
\usepackage{verbatim}
\usepackage{graphicx}
\usepackage{cite}
\usepackage{graphicx}
\usepackage{amsmath}
\usepackage{amssymb}
\usepackage{booktabs}
\usepackage{amsfonts}
\usepackage{array}
\usepackage{hyperref}
\usepackage{comment}

\usepackage{color}
\usepackage{bbding}
\usepackage{pifont}
\usepackage{wasysym}
\usepackage{cite}
\usepackage{stfloats}

\usepackage{multirow}
\usepackage{makecell}
\usepackage{diagbox}
\usepackage{threeparttable}

\usepackage{mathtools}
\usepackage{float}
\hypersetup{
	colorlinks=true,
	linkcolor=black
}

\hypersetup{
	citecolor=black
}
\usepackage{bm}

\DeclareMathOperator{\EX}{\mathbb{E}}

\DeclareMathOperator{\N}{\mathcal{N}}
\DeclareMathOperator{\I}{\mathbf{I}}
\DeclareMathOperator{\x}{\mathbf{x}}

\DeclareMathOperator{\X}{\mathbf{X}}

\def\blfootnote{\xdef\@thefnmark{}\@footnotetext}

\begin{document}
	
	\title{WaveDM: Wavelet-Based Diffusion Models for Image Restoration}
	
	\author{Yi Huang$^*$, Jiancheng Huang$^*$, Jianzhuang Liu,~\IEEEmembership{Senior Member,~IEEE,} \\Mingfu Yan, Yu Dong, Jiaxi Lyu, Chaoqi Chen, Shifeng Chen$^\dagger$
		
		\IEEEcompsocitemizethanks{
			\IEEEcompsocthanksitem $^*$: Equal contribution. $^\dagger$: Corresponding author.
			\IEEEcompsocthanksitem Yi Huang, Jiancheng Huang, Jianzhuang Liu, Mingfu Yan, Yu Dong, Jiaxi Lyu and Shifeng Chen are all with Shenzhen Key Lab of Computer Vision and Pattern Recognition, Shenzhen Institute of Advanced Technology, Chinese Academy of Sciences, Shenzhen, 518055, China and also with University of Chinese Academy of Sciences, Beijing, 100039, China (e-mail: yi.huang; jc.huang; jz.liu; mf.yan; yu.dong; jx.lv1; shifeng.chen@siat.ac.cn).
			\IEEEcompsocthanksitem Chaoqi Chen is with The University of Hong Kong (email: cqchen1994@gmail.com).
		}
		
	}
	
	\markboth{IEEE Transactions on Multimedia}%
	{Shell \MakeLowercase{\textit{et al.}}: A Sample Article Using IEEEtran.cls for IEEE Journals}
	
	\maketitle

	\begin{abstract}
		Latest diffusion-based methods for many image restoration tasks outperform traditional models, but they encounter the long-time inference problem. To tackle it, this paper proposes a Wavelet-Based Diffusion Model (WaveDM).
		WaveDM learns the distribution of clean images in the wavelet domain conditioned on the wavelet spectrum of degraded images after wavelet transform, which is more time-saving in each step of sampling than modeling in the spatial domain. To ensure restoration performance, a unique training strategy is proposed where the low-frequency and high-frequency spectrums are learned using distinct modules.
		In addition, an Efficient Conditional Sampling (ECS) strategy is developed from experiments, which reduces the number of total sampling steps to around 5. Evaluations on twelve benchmark datasets including image raindrop removal, rain steaks removal, dehazing, defocus deblurring, demoiréing, and denoising demonstrate that WaveDM achieves state-of-the-art performance with the efficiency that is comparable to traditional one-pass methods and over 100$\times$ faster than existing image restoration methods using vanilla diffusion models. The code is available at \href{https://github.com/stayalive16/WaveDM}{https://github.com/stayalive16/WaveDM}.
	\end{abstract}
	
	
	\section{Introduction}
	\IEEEPARstart{I}{mage} restoration, aiming to remove degradations (e.g., blur, raindrops, moiré, noise and so on) from a degraded image to generate a high-quality one, has raised great attention in computer vision research. Most previous methods depend on strong priors or estimate the degradation functions for specific tasks \cite{he2010single, kopf2008deep, michaeli2013nonparametric, timofte2013anchored, wang2006real}. With the development of deep learning, deep neural network-driven methods have become the mainstay. These methodologies are primarily built on architectures like Convolutional Neural Networks (CNNs) \cite{anwar2020densely, dudhane2022burst, zamir2020learning, zamir2021multi, jin2019flexible, mou2021cola} and Transformers \cite{liang2021swinir, wang2022uformer, zamir2022restormer, li2023ewt, huang2023learning}. However, some of these deep learning models, generally relying on regression techniques, tend to yield results that are usually over-smoothing and lose subtle details.	On the other hand, unsupervised methods \cite{ma2022toward, jin2019unsupervised,han2020decomposed}, which are implemented without labeled data, promise impressive generalizability, especially in scenarios not seen during training. However, the absence of explicit guidance sometimes results in outputs that may be over-enhanced in colors or contain amplified noise.
	
	Another popular approach is through task-specific generative modeling, frequently leveraging Generative Adversarial Networks (GANs) \cite{qian2018attentive, zhang2019image,li2019heavy, gao2023jdsr, zhang2020supervised}. These generative models aim to capture the latent data distribution of clean images and apply this prior to the degraded samples. While showing powerful generalization capabilities, GAN-based restoration techniques have their own drawbacks. The use of adversarial losses often induces artifacts that are absent in the original clean images, introducing distortions. Besides, the instability of GAN training further intensifies this challenge, and in certain scenarios, can even lead to mode collapse \cite{sohl2015deep}. Another type, flow-based methods \cite{lugmayr2020srflow,whang2021composing}, directly accounts for the ill-posed problem with an invertible encoder, which maps clean images to the flow-space latents conditioned on degraded inputs. However, the need for a strict bijection between latent and data spaces adds to their complexity.
	
	Recently, diffusion models \cite{nichol2021improved, dhariwal2021diffusion, rombach2022high, ho2022cascaded,saharia2022photorealistic} have come into the spotlight. Their achievements span various computer vision tasks such as conditional image generation \cite{rombach2022high, liu2023more, shi2022divae}, image super-resolution \cite{li2022srdiff, saharia2022image}, image-to-image translation \cite{wang2022pretraining, zhao2022egsde, saharia2022palette, sasaki2021unit}, and face restoration \cite{preechakul2022diffusion,yue2022difface, nair2022ddpm}. These models have become popular because of many distinct benefits diffusion models possess. One is the outstanding generative capability as diffusion models can better capture the data distribution compared to other approaches such as GANs. Furthermore, diffusion models excel in countering diverse degradations, ranging from noise and blur to more complex corruptions due to their ability in modeling intricate data distributions. In addition, diffusion models are inherently resistant to mode collapse, ensuring comprehensive data distribution coverage \cite{ho2020denoising, sohl2015deep, song2020improved}. This leads to more stable training, mitigating chances of unpredictable outputs and affirming their reliability in restoration.
	However, the biggest challenge among them is the heavy computational burden as diffusion models usually require many steps of sampling.
	Earlier works \cite{dhariwal2021diffusion, ho2022cascaded} start from generating a low-resolution image and gradually upsample it through pretrained super-resolution models to reduce the processing time in each step. 
	Rombach et al. \cite{rombach2022high} apply the diffusion models in the latent space of powerful pretrained autoencoders for high-resolution image synthesis. Some other works mainly focus on reducing the evaluation steps by accelerated deterministic implicit sampling \cite{song2021denoising}, knowledge distillation\cite{meng2023distillation, salimans2022progressive}, changing the diffusion strategy \cite{lyu2022accelerating} and reformulating the solution to the diffusion ordinary differential equations \cite{lu2022dpm, lu2022dpmplus}. 
	However, they are still restricted to practical applications of high-resolution image restoration.
	
	Recently,  Kawar et al. propose Denoising Diffusion Restoration Models (DDRM) \cite{kawar2022denoising} which takes advantage of a pretrained diffusion model for solving linear inverse restoration problems without extra training, but it cannot handle images with nonlinear degradation.  Some other works \cite{chung2023diffusion, songpseudoinverse, chung2022improving, saharia2022palette, wang2023zero, chung2023parallel, murata2023gibbsddrm} seek to use diffusion models to address nonlinear inverse imaging problems, in which the forms and parameters of the degradation functions have to be known. 
	However, the degradation models of most real-world restoration problems such as deraining cannot be obtained.
	Ozan et al. introduce an approach \cite{ozdenizci2022restoring} to restore vision under adverse weather conditions with size-agnostic inputs by cutting images into multiple overlapping small patches. Although it can process high-resolution images with better performance than traditional one-pass methods, its computational complexity increases quadratically with the increase of image sizes. For example, one $2176\times1536$ image is cut into 12369 $64\times64$ overlapping patches, requiring about 650 seconds for 25-step sampling on a regular GPU.
	
	The first is to decrease the time of processing images in each step. Specifically, WaveDM learns the distribution of clean images in the wavelet domain, which is different from most of the current diffusion models that focus on the spatial domain. 
	After wavelet transform for $n$ times, the spatial size of the original image is reduced by $1/4^n$, thus saving a lot of computation.
	Note that other popular transforms such as Fourier transform cannot achieve this because the size of the Fourier spectrum is the same as that of the image.
	Although some recent works \cite{guth2022wavelet, hui2022neural, phung2023wavelet} also introduce wavelet transform into diffusion models for image or 3D generative tasks, to the best of our knowledge, this attempt has not been explored in addressing the restoration problems.
	In our model, the input images are first decomposed into multiple frequency bands using wavelet transform. In the training phase, a diffusion model is utilized to learn the distribution of low-frequency bands of clean images by perturbing them with random noise at different moments of time. In addition, a lightweight high-frequency refinement module is constructed to provide the high-frequency bands, which also serve as the essential condition. The sampling starts from a random Gaussian noise to predict the low-frequency bands through a reverse diffusion process, which are then combined with the output from the high-frequency refinement module to generate a clean image through inverse wavelet transform.
	
	The second scheme of acceleration is to reduce the total sampling steps, which is realized by an Efficient Conditional Sampling (ECS) strategy we obtain from experiments. ECS follows the same sampling procedure as the deterministic implicit sampling \cite{song2021denoising} during the initial sampling period and then stops at an intermediate moment to predict clean images directly instead of completing all the sampling steps. During this procedure, the degraded images serve as the essential conditions that provide strong priors such as global tone and spatial structure to remove the noise till the end. Due to its simple implementation, ECS can further reduce the sampling steps to as few as 4 without extra computation. Additionally, experimental results on several datasets show that ECS is also capable of maintaining or even improving the restoration performance by setting the intermediate moment reasonably.
	
	The main contributions of this work are summarized as follows:
	
	\begin{itemize}
		\item A wavelet-based diffusion model is proposed to learn the distribution of clean images in the wavelet domain, which dramatically reduces the computational expenses typically encountered in the spatial domain.
		
		\item A unique training strategy is proposed where the low-frequency and high-frequency spectrums are learned using distinct modules, which facilitates the effective restoration of degraded images.
		
		\item An efficient conditional sampling strategy is found based on our experiments to reduce the number of sampling steps to around 5 without extra computation, while maintaining the restoration performance compared to other diffusion-based methods using 25 steps or more.
		
		\item Comprehensive experiments conducted on twelve restoration benchmark datasets verify that our WaveDM achieves state-of-the-art performance with the efficiency that is comparable to traditional one-pass methods and over $100\times$ faster than existing diffusion-based models.

	\end{itemize}
	
	\section{Related Work}
	\label{related work}
	
	\subsection{Image Restoration}
	Earlier restoration methods \cite{he2010single, kopf2008deep, michaeli2013nonparametric, timofte2013anchored} mainly rely on seeking strong priors for specific tasks. In recent years, deep neural works are widely used for general image restoration owing to their superb performance. These learning-based approaches usually require a specific model architecture constructed by CNN \cite{zhang2018image,anwar2020densely, dudhane2022burst, zamir2020learning, zamir2021multi, yi2021structure, zhang2021dual, xin2020wavelet, puthussery2020wdrn,zheng2021learning, liu2020wavelet} or Transformer \cite{liang2021swinir, li2023ewt, wang2022uformer, zamir2022restormer}. Most convolutional encoder-decoder designs \cite{chen2022simple, yi2021efficient, kupyn2019deblurgan, abuolaim2020defocus, cho2021rethinking, li2018multi} could be viewed as variants of a classical solution, U-Net\cite{ronneberger2015u}, the effectiveness of which has been validated for their hierarchical representations while keeping computationally efficient. Extensively, spatial and channel attentions are also injected in it to capture some key information thus boosting the performance. 
	The Vision Transformer \cite{dosovitskiy2021image, liu2021swin}, first introduced for image classification, is capable of building strong relationships between image patches due to its self-attention mechanism. Naturally, a lot of transformer-based works are also studied for the low-level vision tasks like super-resolution \cite{yang2020learning, liang2021swinir,lin2023steformer}, denoising \cite{wang2022uformer, li2023ewt, zamir2022restormer}, deraining \cite{xiao2022image}, colorization \cite{kumar2021colorization}, etc. Different from them, this paper aims to tackle the restoration problem from the view of generative modeling, implemented by a wavelet-based diffusion model.
	
	\subsection{Diffusion Models}
	Diffusion models, a new type of generative models, are inspired by non-equilibrium thermodynamics. They learn to reverse the forward process of sequentially corrupting data samples with additive random noise following the Markov chain, until reconstructing the desired data that matches the source data distribution from noise. Previous diffusion models can be roughly classified into diffusion based \cite{sohl2015deep} and score-matching based \cite{hyvarinen2005estimation, vincent2011connection}. Following them, denoising diffusion probabilistic models \cite{ho2020denoising, nichol2021improved} and noise-conditioned score network \cite{song2019generative, song2021score, song2020improved} are proposed to synthesize high-quality images, respectively.
	
	\subsubsection{Diffusion Models in Low-Level Vision Tasks}Recently, diffusion-based models show great potential in various computer vision tasks under conditions such as class-conditioned image synthesis with and without classifier guidance \cite{dhariwal2021diffusion, kawar2022enhancing, ho2021classifier}, image inpainting \cite{lugmayr2022repaint}, super-resolution \cite{lee2022progressive, saharia2022image}, deblurring \cite{whang2022deblurring}, and image-to-image translation (e.g., colorization and style transfer) \cite{wang2022pretraining, saharia2022palette, choi2021ilvr, kwon2023diffusion, zhang2023inversion}. Similarly, the applications following the score-based conditional modeling are also widely explored \cite{meng2022sdedit, chung2022come}.
	Beyond synthesis, some works apply diffusion models for image restoration. Most of the restoration methods are trained either on large-scale datasets or with samples that come from some specific types (e.g., faces \cite{preechakul2022diffusion, nair2022ddpm}) to obtain high-quality generation performance. However, they may somewhat change the original spatial structure of conditional degraded images. Kawar et al. \cite{kawar2022denoising} propose DDRM to solve linear inverse image restoration problems, but it cannot be adapted to the inversion of nonlinear degradation. Some works \cite{chung2023diffusion, songpseudoinverse, chung2022improving, saharia2022palette, wang2023zero, chung2023parallel, murata2023gibbsddrm} use diffusion models to address nonlinear inverse imaging problems, in which the forms and parameters of the degradation functions have be to known. Ozan et al. \cite{ozdenizci2022restoring} propose patch-based diffusion models, which is the first diffusion-based work that achieves state-of-the-art performance on three real-world blind restoration tasks in terms of pixel-wise evaluation metrics such as PSNR. However, the main limitation of it is the much longer inference time than traditional one-pass methods due to a large amount of image patches and many sampling steps. This paper aims to solve this problem through the wavelet-based diffusion model with an efficient conditional sampling strategy, preserving the state-of-the-art restoration performance simultaneously.
	
	\subsubsection{Accelerating Diffusion Models}
	Though diffusion models are capable of generating high-quality images, their iterative sampling procedure usually results in long inference time. Song et al. \cite{song2021denoising} propose the deterministic implicit sampling that requires only 25 steps. Lu et al. \cite{lu2022dpm} reformulate the exact solution to the diffusion ordinary differential equations (ODEs) and propose a fast dedicated high-order solver for diffusion ODE speedup using around 10 steps. Ma et al. \cite{ma2022accelerating} investigate this problem by viewing the diffusion sampling process as a Metropolis adjusted Langevin algorithm  and introduce a model-agnostic preconditioned diffusion sampling that leverages matrix preconditioning, which accelerates vanilla diffusion models by up to $29\times$. Lyu et al. \cite{lyu2022accelerating} start the reverse denoising process from a non-Gaussian distribution, which enables stopping the diffusion process early where only a few initial diffusion steps are considered. However, it requires an extra generative model (e.g., GAN or VAE) to approximate the real data distribution to start sampling. Different from them, our work focuses on accelerating the conditional image restoration diffusion model from two aspects: reducing the processing of each step implemented by a wavelet-based diffusion model, and reducing the number of total sampling steps by an efficient conditional sampling strategy without extra training.
	
	\subsection{Wavelet Transform-Based Methods}
	Wavelet transform has been widely explored in computer vision tasks, especially combined with deep neural networks. For example, Liu et al. \cite{liu2018multi} propose a multilevel Wavelet-CNN to enlarge receptive fields with a better trade-off between efficiency and restoration performance via multi-level wavelets. Liu et al. \cite{liu2020wavelet} design a wavelet-based dual-branch network with a spatial attention mechanism for image demoiréing. 
	Xin et al. \cite{xin2020wavelet} first decompose the low-resolution image into a series of wavelet coefficients (WCs) and then use a CNN to predict the corresponding series of high-resolution WCs, which are then utilized to reconstruct the high-resolution image.
	Li et al. \cite{li2023ewt} propose an efficient wavelet transformer for image denoising. It is the first attempt to utilize Transformer in the wavelet domain, implemented by an efficient multi-level feature aggregation module, thus significantly reducing the device resource consumption of the conventional Transformer model. 
	All the methods mentioned above combine wavelet transform with deep neural networks like CNNs and Transformers by designing task-specific network structures without using diffusion models, while our method combines a diffusion model with wavelet transform for various image restoration tasks by employing the general convolutional U-Net architecture.
	Our approach can achieve superior performance on multiple image restoration tasks while maintaining comparable processing efficiency.
	
	In recent years, wavelet diffusion-based methods have emerged as a prominent approach, particularly in the realm of image generation. As evident from the works of Phung et al. \cite{phung2023wavelet} and Guth et al. \cite{guth2022wavelet}, they focus on leveraging wavelet diffusion for image synthesis. Hui et al. \cite{hui2022neural} extend the framework to 3D shape generation. These approaches, while significant, primarily target generation tasks. Different from them, our WaveDM is architected with deliberate design, leveraging the diffusion principle innovatively for image restoration.

	\section{Preliminaries}
	\label{preliminaries}

	\subsection{Denoising Diffusion Probabilistic Models}
	
	Denoising Diffusion Probabilistic Models (DDPMs) \cite{ho2020denoising, nichol2021improved} are a class of generative models that work by destroying training data through the successive addition of Gaussian noise, and then learning to recover the data by reversing this noising process. During training, the forward noising process follows the Markov chain that transforms a data sample from the real data distribution $\x_0\sim q(\x_0)$ into a sequence of noisy samples $\x_t$ in $T$ steps with a variance schedule $\beta_1,\ldots,\beta_T$:
	\begin{equation}
		q(\x_{t}\vert\x_{t-1}) = \mathcal{N}(\x_{t};\sqrt{1-\beta_t}\x_{t-1},\beta_t \I),
		\label{eq:forward}
	\end{equation}
	
	Diffusion models learn to reverse the above process through a joint distribution $p_{\theta}(\x_{0:T})$ that follows the Markov chain with parameters $\theta$, starting at a noisy sample from a standard Gaussian distribution $p(\x_T)=\mathcal{N}(\x_T;\mathbf{0},\I)$:
	\begin{equation}
		p_{\theta}(\x_{0:T}) = p(\x_{T}) \prod_{t=1}^T p_{\theta}(\x_{t-1}\vert\x_t),
	\end{equation}
	\begin{equation}
		p_{\theta}(\x_{t-1}\vert\x_t) = \mathcal{N}(\x_{t-1};\bm{\mu}_{\theta}(\x_t,t),\mathbf{\Sigma}_{\theta}(\x_t,t)).
		\label{eq:reverse}
	\end{equation}
	The parameters $\theta$ are usually optimized by a neural network that predicts $\bm{\mu}_{\theta}(\x_t,t)$ and $\mathbf{\Sigma}_{\theta}(\x_t,t)$ of Gaussian distributions, which is simplified by predicting noise vectors $\bm{\epsilon}_{\theta}(\x_t,t)$ with the following objective \cite{ho2020denoising}:
	\begin{center}
		\begin{equation}
			\begin{aligned}
				\mathbb{E}_{q(\x_0)}[-\log p_{\theta}(\x_0)]\leq \EX_{q}\Big[-\log \frac{p_{\theta}(\x_{0:T})}{q(\x_{1:T}\vert\x_0)}\Big] \\
				= \EX_{q} \Big[ \underbrace{D_{\text{KL}}(q(\x_T|\x_0)\,||\,p(\x_T))}_{L_{T}} \underbrace{-\log p_{\theta}(\x_0|\x_1)}_{L_0} \\ 
				+ \sum_{t>1}\underbrace{D_{\text{KL}}(q(\x_{t-1}|\x_t,\x_0)\,||\,p_{\theta}(\x_{t-1}|\x_t))}_{L_{t-1}} \Big].
				\label{eq:obj_expanded}
			\end{aligned}
		\end{equation}
	\end{center}
	Obviously, the $L_{t-1}$ term actually trains the network to perform one reverse diffusion step. As reported by \cite{ho2020denoising}, the optimization of $L_{t-1}$ can be converted to training a network $\bm{\mu}_{\theta}(\x_t,t)$ that estimates the mean value of the posterior distribution $q(\x_{t-1}|\x_t,\x_0)$. Furthermore, the model can instead be trained to predict the noise vector $\bm{\epsilon}_{\theta}(\x_t,t)$ using an alternative reparameterization of the reverse process by:
	\begin{equation}
		\bm{\mu}_{\theta}(\x_t,t) = \frac{1}{\sqrt{\alpha_t}} \left(\x_t - \frac{\beta_t}{\sqrt{1-\bar{\alpha}_t}}\bm{\epsilon}_{\theta}(\x_t,t)\right),
		\label{eq:reparameterization}
	\end{equation}
	where $\alpha_t=1-\beta_t$, $\bar{\alpha}_t=\prod_{i=1}^t\alpha_i$.
	As a result, the training objective is transformed into a re-weighted simplified form given as:
	\begin{equation}
		L_{simple} =  \mathbb{E}_{\x_0,t,\bm{\epsilon}_t\sim\N(\mathbf{0},\I)}\Big[\vert\vert\bm{\epsilon}_t -  \bm{\epsilon}_{\theta}(\x_t,t)\vert\vert^2 \Big].
		\label{eq:training_obj}
	\end{equation}
	Consequently, the sampling phase with the learned parameterized Gaussian transitions $p_{\theta}(\x_{t-1}\vert\x_t)$ can start from $\x_T\sim\N(\mathbf{0},\I)$ by:
	\begin{equation}
		\x_{t-1}=\frac{1}{\sqrt{\alpha_t}}\left(\x_t - \frac{\beta_t}{\sqrt{1-\bar{\alpha}_t}} \bm{\epsilon}_{\theta}(\x_t,t)\right) + \sigma_t\bm{z},
	\end{equation}
	where $\bm{z}\sim\mathcal{N}(\mathbf{0},\I)$, $\alpha_t=1-\beta_t$, and $\bar{\alpha}_t=\prod_{i=1}^t\alpha_i$.
	
	\subsection{Deterministic Implicit Sampling}\label{DDIM sampling}
	
	Denoising Diffusion Implicit Models (DDIMs) \cite{song2021denoising} generalize DDPMs to obtain the same training objective as Eq. \ref{eq:training_obj} by defining a non-Markovian diffusion process:
	\begin{equation}
		q_{\sigma}(\x_{t-1}\vert\x_t,\x_0)=\N(\x_{t-1};\bm{\Tilde{\mu}}_t(\x_t,\x_0),{\sigma}_t^2\I).
		\label{ddim sampling}
	\end{equation}
	By setting $\sigma_t^2=\frac{1-\bar{\alpha}_{t-1}}{1-\bar{\alpha}_t}\beta_t$, the forward process becomes Markovian and remains the same as DDPMs.
	
	A deterministic implicit sampling (also called DDIM sampling) is implemented by setting $\sigma_t^2=0$, and thus the sampling process based on Eq. \ref{ddim sampling} can be accomplished by: 
	\begin{equation}
		\begin{split}
			\x_{t-1} &=  \sqrt{\bar{\alpha}_{t-1}}\left(\frac{\x_t-\sqrt{1-\bar{\alpha}_t}\cdot\bm{\epsilon}_{\theta}(\x_t,t)}{\sqrt{\bar{\alpha}_t}}\right) \\ & + \sqrt{1-\bar{\alpha}_{t-1}}\cdot\bm{\epsilon}_{\theta}(\x_t,t),
		\end{split}
		\label{eq:ddim}
	\end{equation}
	which enables a faster sampling procedure. Specifically, DDIMs replace the complete reverse sampling sequence ${\x_T,\x_{T-1},\ldots,\x_1,\x_0}$ with one of its sub-sequence $\x_T, \x_{\tau_S},\x_{\tau_{S-1}},\ldots,\x_{\tau_1}$ which can be obtained by:
	\begin{equation}
		\tau_i = (i-1)\cdot T / S,
		\label{acceleration}
	\end{equation}
	where $S$ denotes the total sampling steps for acceleration. Thus, the faster DDIM sampling  procedure is formulated as:
	
	\begin{equation}
		\begin{split}
			\x_{t-1} & =  \sqrt{\bar{\alpha}_{t-1}}\left(\frac{\x_t-\sqrt{1-\bar{\alpha}_t}\cdot\bm{\epsilon}_{\theta}(\x_t,t)}{\sqrt{\bar{\alpha}_t}}\right) \\ &  + \sqrt{1-\bar{\alpha}_{t-1}}\cdot\bm{\epsilon}_{\theta}(\x_t,t), \  \  t = T, \tau_S, \ldots, \tau_1.
		\end{split}
		\label{eq_ddim_faster}
	\end{equation}

	\subsection{Wavelet Transform}
	
	\subsubsection{2D Discrete Wavelet Transform (2D DWT)}
	
	Given an image \( I \in \mathbb{R}^{H\times W\times C} \), where \( M \times N \) is the spatial size, $C$ is the number of channesl, the 2D DWT decomposes the image into four sub-bands:
	\begin{equation}
		I_{LL}, I_{LH}, I_{HL}, I_{HH} = \text{DWT}_{2D}(I).
	\end{equation}
	The sub-band \( I_{LL} \) represents the approximation coefficients and has a size of \( \frac{M}{2} \times \frac{N}{2} \times C \). The other sub-bands \( I_{LH} \), \( I_{HL} \), and \( I_{HH} \) correspond to the horizontal, vertical, and diagonal detail coefficients, respectively, and each of them possesses a size of \( \frac{M}{2} \times \frac{N}{2} \times C \).
	
	For multi-level wavelet decomposition, the DWT is recursively applied to the \( I_{LL} \) sub-band from the previous level. After \( k \) decompositions, the size of the \( I_{LL} \) sub-band reduces to \( \frac{M}{2^k} \times \frac{N}{2^k} \times C \).
	
	The wavelet used for decomposition can be of various types, such as the Haar wavelet, which provides a simple and effective basis for image decomposition.
	
	\subsubsection{2D Inverse Discrete Wavelet Transform (2D IDWT)}

	Starting with the four sub-bands, the original image is reconstructed:
	\begin{equation}
		I' = \text{IDWT}_{2D}(I_{LL}, I_{LH}, I_{HL}, I_{HH}),
	\end{equation}
	where \( I' \in \mathbb{R}^{H\times W\times C} \). The process of multi-level reconstruction begins from the deepest decomposition level and sequentially moves towards the first level, eventually yielding a reconstructed image \( I' \) of the original size.
	
	\subsubsection{2D Full Wavelet Packet Transform (2D FWPT)}
	
	Unlike the 2D DWT, which only recursively decomposes the \( I_{LL} \) sub-band, the 2D Full Wavelet Packet Transform (2D FWPT) exhaustively decomposes every sub-band at each level.
	
	For a single level decomposition of an image \( I \in \mathbb{R}^{H\times W\times C} \), the 2D FWPT yields 4 sub-bands:
	\begin{equation}
		\{I_{i,j}\}_{i,j \in \{L, H\}} = \text{FWPT}_{2D}(I),
	\end{equation}
	where each sub-band \( I_{i,j} \in \mathbb{R}^{\frac{M}{2} \times \frac{N}{2} \times C} \).
	
	For a 2-level FWPT, each of the initial sub-bands is further decomposed, leading to a total of 16 sub-bands, each of size \( \frac{M}{4} \times \frac{N}{4} \times C \).
	
	The benefit of this exhaustive decomposition is that all the sub-bands at each level have the same spatial dimension, allowing easier concatenation and analysis of frequency details in a structured manner.
	
	\subsubsection{2D Inverse Full Wavelet Packet Transform (2D IFWPT)}
	
	Given the sub-bands obtained from the 2D FWPT, the 2D IFWPT reconstructs the original image. For a 1-level FWPT:
	\begin{equation}
		I' = \text{IFWPT}_{2D}(\{I_{i,j}\}_{i,j \in \{L, H\}}),
	\end{equation}
	where the reconstructed image \( I' \in \mathbb{R}^{H\times W\times 1} \). The reconstruction process, similar to 2D IDWT, starts from the deepest decomposition level and works its way up to the first level, combining all sub-bands together to form the original image.

	\begin{figure*}[t]
		\small
		\centering
		\includegraphics[width=\textwidth]{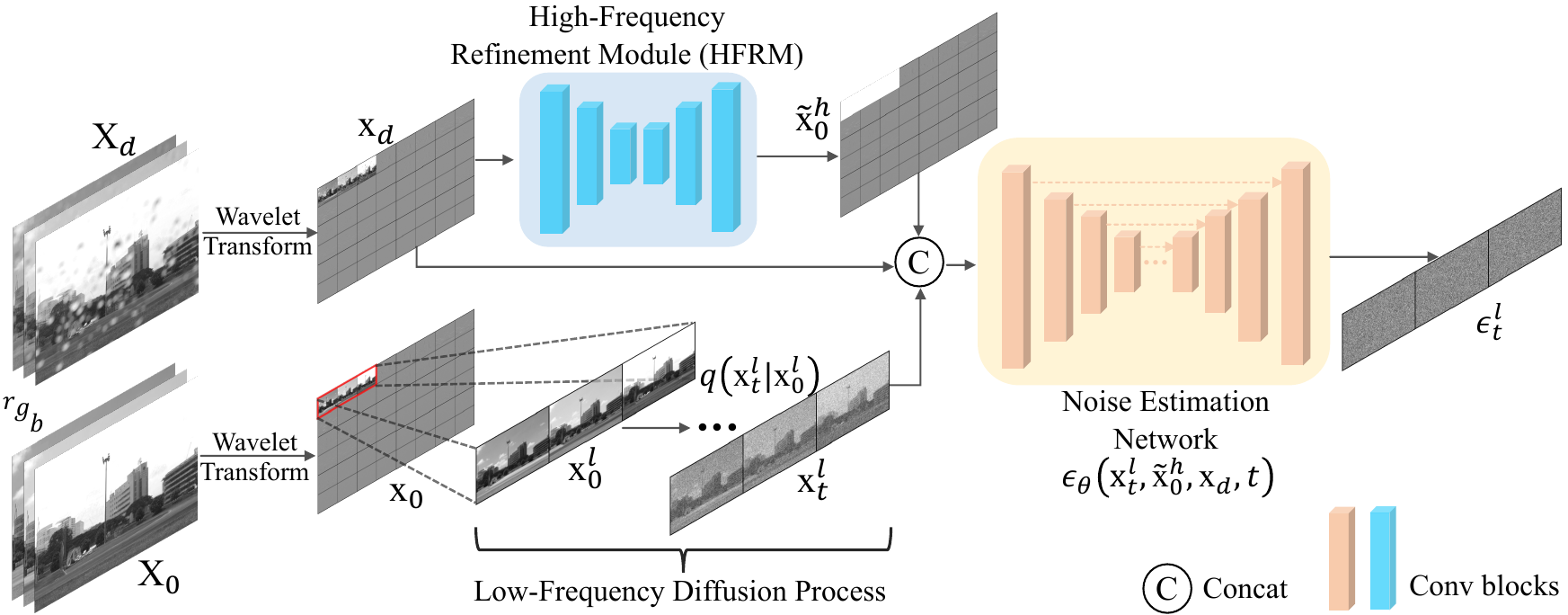}
		\caption{Training of the wavelet-based diffusion model (WaveDM) for image restoration, where $\X_d$ and $\X_0$ stand for a pair of RGB degraded and clean images. $\x_d$ and $\x_0$ are the wavelet spectrum of $\X_d$ and $\X_0$ after the Haar wavelet transform, respectively. $\x_t^l$ is the diffusion result of the low-frequency spectrum $\x_0^l$ extracted from the first three bands of $\x_0$. $\tilde{\x}_0^h$ denotes the high-frequency spectrum of the clean image based on $\x_d$ with the HFRM. $\x_d$, $\tilde{\x}_0^h$ and $\x_t^l$ are concatenated together as input to the noise estimation network $\bm{\epsilon}_{\theta}(\x_t^l,\tilde{\x}_0^h,\x_d,t)$ to predict the noise $\bm{\epsilon}_t^l$ at all time moments.}
		\label{train_diff}
	\end{figure*}
	
	\section{Method}

	\subsection{Overview}
	Recently, diffusion models are increasingly favored over alternatives like GANs in image restoration due to their better ability to capture complex data distributions and their inherently stable training processes. However, current methods, such as \cite{ozdenizci2022restoring} and \cite{kawar2022denoising}, apply diffusion models directly in the spatial domain, resulting in long inference time. To mitigate this computational challenge, we leverage the wavelet transform's capability for image size reduction with no information loss and frequency sub-band separation. Consequently, we propose a wavelet-based diffusion model (WaveDM) that learns the distribution of clean images in the wavelet domain, where the low-frequency and high-frequency spectrums are learned using distinct modules for restoration quality.

	Fig.~\ref{train_diff} depicts the WaveDM training procedure. The pipeline consists of three primary parts: the High-Frequency Refinement Module (HFRM), the low-frequency diffusion process, and the noise estimation network. Firstly, both degraded and clean image pairs are transformed to the wavelet domain using the 2D FWPT, with the high and low-frequency details extracted and resolution reduced. 
	The full wavelet spectrum of the degraded image is taken as input to HFRM to estimate the clean image's high-frequency spectrum. Concurrently, we add Gaussian noise to the low-frequency spectrum of the clean image. The noisy low-frequency spectrum, concatenated with the input and output of HFRM, is then sent to the noise estimation network for noise prediction. Comprehensive training details are described in Section~\ref{trainingofwavedm}.

\begin{figure}[t]
		\centering
		\includegraphics[width=0.5\textwidth]{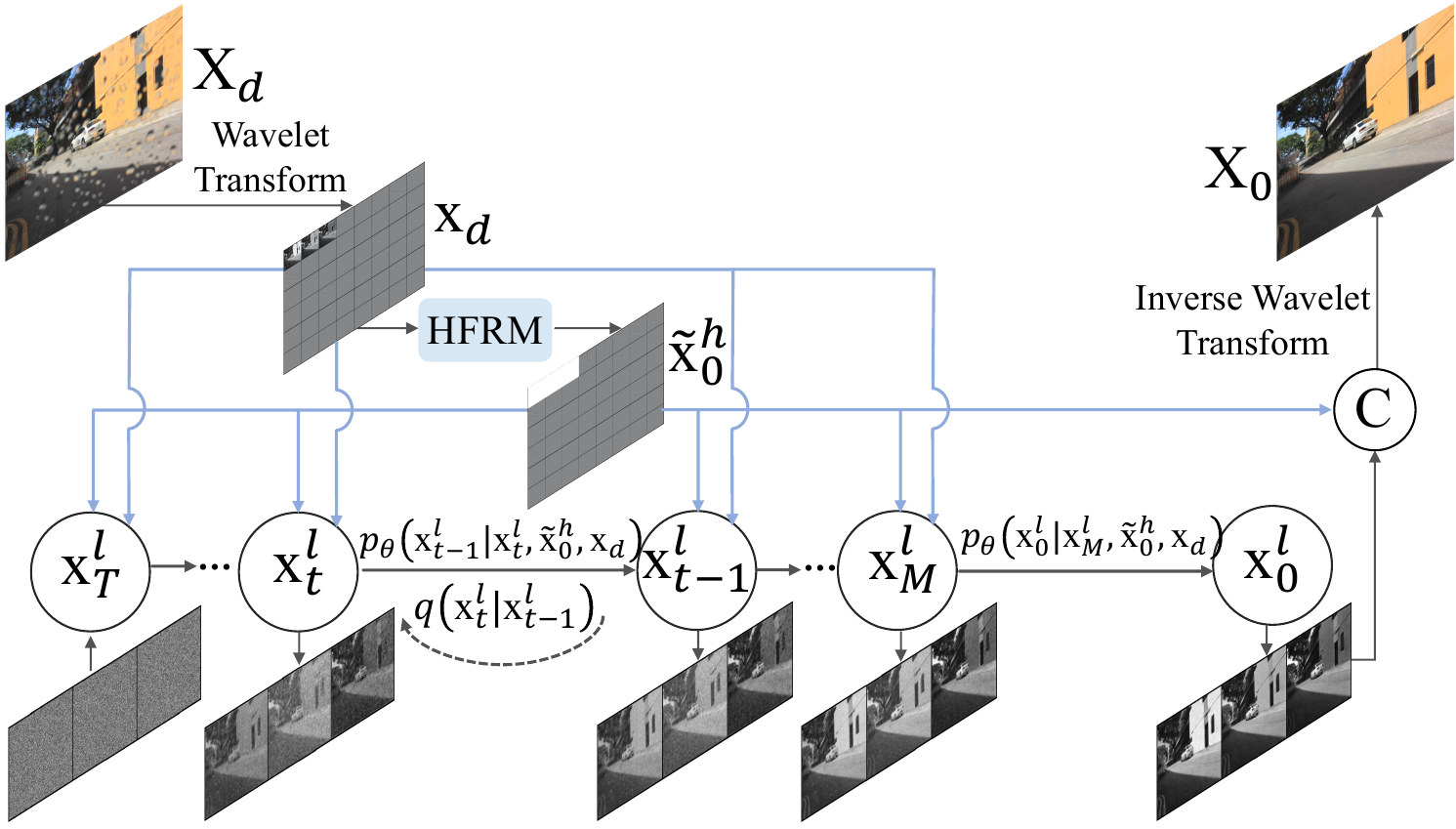}
		\caption{Overview of WaveDM with ECS. $q(\x_t^l\vert\x_{t-1}^l)$ stands for the forward diffusion (dashed line). The sampling process $p_{\theta}(\x_{t-1}^l\vert\x_t^l, \tilde{\x}_0^h, \x_d)$ (solid lines) starts from a standard Gaussian noise $\x_T^l \sim\mathcal{N}(\mathbf{0},\I)$ to generate the low-frequency spectrum of the clean image, where $\x_d$ and $\tilde{\x}_0^h$ serve as conditions (blue solid lines) from step $T$ to step $M$. Then the intermediate result $\x_M^l$ is utilized to predict the low-frequency spectrum $\x_0^l$ of the clean image directly, followed by inverse wavelet transform that turns the concatenation of $\tilde{\x}_0^h$ and $\x_0^l$ into a clean RGB image $\X_0$.}
		\label{sample_diff}
	\end{figure}

	Fig.~\ref{sample_diff} describes the sampling process of WaveDM. First, 2D FWPT captures the wavelet spectrum of a degraded image, serving as HFRM's input. The sampling starts from the concatenation of HFRM's input and output with a Gaussian noise, which is then sent to the noise estimation network, yielding a noisy low-frequency wavelet spectrum at the first step. This process iterates using the efficient conditional sampling strategy, described in Section~\ref{samplingingofwavedm}, to produce a clean low-frequency wavelet spectrum at the end of sampling. Then the final clean RGB image is obtained from the concatenation of this spectrum with HFRM's output, followed by 2D IFWPT.

	\subsection{Training of WaveDM}
	\label{trainingofwavedm}

	As shown in Fig.~\ref{train_diff}, given a degraded image \( \mathbf{X}_d \in \mathbb{R}^{H\times W\times 3} \) and its corresponding ground truth \( \mathbf{X}_0 \in \mathbb{R}^{H\times W\times 3} \), we employ a 2-level 2D FWPT using the Haar wavelet. The Haar wavelet transform iteratively applies low-pass and high-pass decomposition filters, coupled with downsampling, to compute the wavelet coefficients. Specifically, the low-pass filter, with coefficients \(( \frac{1}{\sqrt{2}},  \frac{1}{\sqrt{2}} )\), captures the average information, while the high-pass filter, with coefficients \(( \frac{1}{\sqrt{2}}, -\frac{1}{\sqrt{2}} )\), focuses on the details or transitions in the image. The transformation process begins by applying these filters to each row of the image, resulting in two intermediate forms. These forms are then subjected to the same filter application along their columns, decomposing the original image into four distinct sub-bands: LL (averaged information), LH (details along columns), HL (details along rows), and HH (details in both rows and columns). For the 2-level 2D FWPT, this decomposition process is recursively applied to all sub-bands. As a result, each image is transformed into the wavelet spectrum \( \mathbf{x}_d, \mathbf{x}_0 \in \mathbb{R}^{\frac{H}{4}\times \frac{W}{4}\times 48} \), consisting of 48 bands with the same spatial dimension, which can be represented as:
	\begin{equation}
		\begin{aligned}
			\x_d = \text{FWPT}_{2D}(\X_d),\\
			\x_0 = \text{FWPT}_{2D}(\X_0).
			\label{eq_wt}
		\end{aligned}
	\end{equation}
	
	Instead of adopting a naive diffusion approach in the wavelet domain, which directly corrupts all wavelet bands of the clean image using additive Gaussian noise and then reversing the process during sampling, we introduce an optimized approach. Essential experiments, discussed in Section~\ref{bandchoiceanalysis}, demonstrated the ineffectiveness of the naive method. Drawing from experimental insights, only the low-frequency spectrum $\x_0^l \in \mathbb{R}^{\frac{H}{4}\times \frac{W}{4}\times 3}$ which is derived from the first three bands of the clean image wavelet spectrum $\x_0$, is corrupted with Gaussian random noise. This corruption follows a forward diffusion process defined as:
	$q(\x_t^l\vert\x_0^l)=\mathcal{N}(\x_t^l;\sqrt{\bar{\alpha}_t}\x_0^l,(1-\bar{\alpha}_t)\I), t={1,2,\ldots,T}$.
	
	Additionally, recognizing the importance of high-frequency information that remains unmodeled in the low-frequency spectrum, we design a High Frequency Refinement Module (HFRM). This lightweight module estimates the high-frequency spectrum $\tilde{\x}_0^h\in \mathbb{R}^{\frac{H}{4}\times \frac{W}{4}\times 45}$ of the clean image $\x_0$ from $\x_d$ in a single pass. This can be presented as:
	\begin{equation}
		\begin{aligned}
			\tilde{\x}_0^h = \text{HFRM}(\x_d).\\
		\end{aligned}
		\label{eq_hfrm}
	\end{equation}
	
	During each step, the degraded image's wavelet spectrum $\x_d$ and the estimated high-frequency spectrum $\tilde{\x}_0^h$ serve as conditions to model the low-frequency spectrum distribution of clean images. Specifically, the concatenated diffusion result $\x_t^l, t=1,2,\ldots,T$, $\tilde{\x}_0^h$, and $\x_d$ across channels feed into the noise estimation network $\bm{\epsilon}_{\theta}(\x_t^l,\tilde{\x}_0^h,\x_d,t)$. By transitioning the diffusion model from the spatial domain to the wavelet domain using 2D FWPT, we achieve a spatial size reduction of $1/16$ for input images, leading to a substantial speedup in processing.
	
	For training, we employ a combined objective function to optimize the diffusion process in the wavelet domain and refine the high-frequency bands. Specifically, the primary objective $L_{simple}$, as defined in Eq.~\ref{eq:training_obj}, is utilized to optimize $\bm{\epsilon}_{\theta}$. The HFRM, which is independent of the variable $t$, is trained using the objective $L_1=\vert\vert\tilde{\x}_0^h-\x_0^h\vert\vert_1$, where $\x_0^h$ denotes the high-frequency bands of $\x_0$. The total training loss is given by:
	\begin{equation}
		\label{total_loss}
		L_{total} = L_{simple} + \lambda L_1.
	\end{equation}
	where $\lambda$ acts as a weighting hyperparameter.

	\label{samplingingofwavedm}
	
	\subsection{Sampling of WaveDM}
	The WaveDM framework, after training, adopts a sequential inference approach in processing the wavelet bands. Firstly, high-frequency wavelet bands are predicted. Subsequently, the low-frequency wavelet bands are sampled. These combined bands are then utilized to generate a clean RGB image using 2D IFWPT. This entire operation is represented in Fig.~\ref{sample_diff}.
	
	For a degraded image denoted by \(\X_d \in \mathbb{R}^{H\times W\times 3}\), we apply a 2-level 2D FWPT implemented by the Haar wavelet. This transformation uses the same filters as in the training of WaveDM. The output of this operation is the wavelet spectrum \( \mathbf{x}_d \in \mathbb{R}^{\frac{H}{4}\times \frac{W}{4}\times 48}\) as outlined in Eq.~\ref{eq_wt}. Then the spectrum \(\x_d\), when fed into HFRM, produces the predicted high-frequency bands \(\tilde{\x}_0^h\) of the restored image, which is described in Eq.~\ref{eq_hfrm}.
	
	To estimate the low-frequency wavelet bands of the restored image, both \(\x_d\) and \(\tilde{\x}_0^h\) are employed. This operation conventionally begins with a random Gaussian noise generation, denoted as \(\x_T^l\in \mathbb{R}^{\frac{H}{4}\times \frac{W}{4}\times 3} \sim\mathcal{N}(\mathbf{0},\I)\) at timestep $T$. Usually, this noise acts as a starting point for the DDIM sampling which samples a clean low-frequency spectrum \(\x_0^l\). The procedure using DDIM sampling is methodically detailed in Eq.~\ref{eq_ddim_wavedm}:
	\begin{equation}
		\begin{split}
			\x_{t-1}^l &=  \sqrt{\bar{\alpha}_{t-1}}\left(\frac{\x_t^l-\sqrt{1-\bar{\alpha}_t}\cdot\bm{\epsilon}_{\theta}(\x_t^l,\x_d, \tilde{\x}_0^h,t)}{\sqrt{\bar{\alpha}_t}}\right) \\ & + \sqrt{1-\bar{\alpha}_{t-1}}\cdot\bm{\epsilon}_{\theta}(\x_t^l,\x_d, \tilde{\x}_0^h,t), \  \  t = T, \tau_S, \ldots, \tau_1,
		\end{split}
		\label{eq_ddim_wavedm}
	\end{equation}
	where the number of total sampling steps is $S$.
	
	However, in our experimental observations, using DDIM sampling directly requires more than 20 steps for achieving the desired restoration performance. This inefficiency in DDIM sampling drives us to explore alternative strategies. After conducting extensive experiments, we find and develop the Efficient Conditional Sampling (ECS) strategy, the effectiveness and efficiency of which are demonstrated in Section~\ref{exp_ecs}. Not only does ECS significantly reduce the sampling steps to around 5, but also brings an enhancement in the restoration quality compared to the conventional DDIM sampling.
	
	In the ECS methodology, instead of allowing DDIM sampling to run its full process, we strategically interrupt it at a specific intermediate step denoted as \(M\). At this moment, rather than continuing with the usual diffusion sampling iterations, we leverage the information contained in the noisy spectrum \(\x_M^l\). With this information, we compute the desired \(\x_0^l\) directly by a portion of the DDIM equation (Eq.~\ref{eq_ddim_wavedm}), effectively simplifying the process and mitigating the need for additional iterative steps. This ECS procedure is represented in Eq.~\ref{eq_ecs_wavedm}, in which the number of total sampling steps is $S(T-M)/T +1$. It is noteworthy that for \(\bm{\epsilon_{\theta}}\), the input variables \(\x_t^l,\x_d,\tilde{\x}_0^h\) are concatenated channel-wise.
	\begin{equation}
		\begin{aligned}
			\begin{cases}
				\begin{split}
					\x_{t-1}^l &= \sqrt{\bar{\alpha}_{t-1}}\left(\frac{\x_t^l-\sqrt{1-\bar{\alpha}_t}\cdot\bm{\epsilon}_{\theta}(\x_t^l,\x_d,\tilde{\x}_0^h,t)}{\sqrt{\bar{\alpha}_t}}\right) \\ & + \sqrt{1-\bar{\alpha}_{t-1}}\cdot\bm{\epsilon}_{\theta}(\x_t^l,\x_d,\tilde{\x}_0^h,t), \  t = T, \tau_S, \ldots, M+\frac{T}{S},
				\end{split}
				\\\\
				\begin{split}
					\ \ \hat{\x}_0^l \; = \; \frac{\x_M^l-\sqrt{1-\bar{\alpha}_M}\cdot\bm{\epsilon}_{\theta}(\x_M^l,\x_d,\tilde{\x}_0^h,M)}{\sqrt{\bar{\alpha}_M}}.
				\end{split}
			\end{cases} 
		\end{aligned}
		\label{eq_ecs_wavedm}
	\end{equation}

	Upon acquiring the clean low-frequency wavelet spectrum \(\hat{\x}_0^l\) through Eq.~\ref{eq_ecs_wavedm}, the restored clean RGB image \(\X_0\) is obtained using 2D IFWPT. This is expressed as:
	\begin{equation}
		\begin{aligned}
			\X_0 = \text{IFWPT}_{2D}(\hat{\x}_0^l, \tilde{\x}_0^h), \\
		\end{aligned}
		\label{eq_iwt}
	\end{equation}
	where \(\tilde{\x}_0^h\) is the high-frequency wavelet spectrum predcited by HFRM.

	\section{Experiments}
	\label{experiments}
	\subsection{Datasets and Settings}
	
	We evaluate WaveDM on twelve benchmark datasets for several image restoration tasks: (i) RainDrop \cite{qian2018attentive} (861 training images and 58 testing images of size $720\times480$) for image raindrop removal, (ii) Outdoor-rain \cite{li2019heavy} (9000 training images and 750 testing images of size $720\times480$) for image rain steaks removal, (iii) SOTS-Outdoor\cite{li2018benchmarking} (72135 training images and 500 testing images with about $600\times400$ resolution) for image dehazing, (iv) DPDD \cite{abuolaim2020defocus} (350 training images of size $1680\times1120$ and 76 testing images of size $1664\times1120$) for both single-pixel and dual-pixel defocus deblurring, (v) London's Buildings \cite{liu2020wavelet} (561 training images and 53 testing images with about $2200\times1600$ resolution) for image demoiréing, (vi) SIDD\cite{abdelhamed2018high} (about 30000 training images and 1280 testing images of size $256\times256$) for real image denoising, and (vii) DFWB for training with 6 benchmark datasets for testing Gaussian image denoising. Specifically, DFWB denotes the combination of DIV2K\cite{agustsson2017_CVPR_Workshops} (800 images), Flickr2K\cite{lim2017enhanced} (2650 images), WED \cite{ma2016waterloo} (4744 images), and BSD500 \cite{martin2001database} (400 images).
	
	The framework of the Patch-based Diffusion Models \cite{ozdenizci2022restoring} (PatchDM) is adopted as the baseline, with which we share the same training settings.
	(e.g., 1000 diffusion steps with linear noise corruption strategy, sinusoidal positional encoding\cite{vaswani2017attention} to encode time embeddings for $t$, 2000000 training iterations, Adam optimized with a fixed learning rate of $4\times e^{-4}$ without weight decay, and exponential moving average with a weight of 0.999 to facilitate more stable training). 
	A similar U-Net architecture based on WideResNet \cite{zagoruyko2016wide} is used as the backbone of the noise estimation network with minor revision to adapt to the input size. As for HFRM, we use the same architecture with fewer residual blocks and reduced number of feature channels. The whole training is implemented on eight NVIDIA Tesla V100 GPUs. 
	We use Peak Signal-to-Noise Ratio (PSNR), Structural Similarity Index Measure (SSIM), Frechet Inception Distance (FID), and inference time on a single NVIDIA Tesla V100 GPU as the main evaluation metrics. Besides, we also test the number of model parameters and memory consumption for reference.
	
	\begin{table}[t]
		\small
		\renewcommand\arraystretch{1.1}
		\begin{center}
			\caption{Comparison of different settings for learning the distributions of clean images on RainDrop. $Comp.:$ diffusion components. $Cond.:$ conditional components. \CheckmarkBold: used. \XSolidBrush: not used.
			}\label{table1}
			\resizebox{0.95\linewidth}{!}{
				\begin{tabular}{c|c|c|c|c|c|c}
					\toprule
					Method&HFRM&$Comp.$&$Cond.$&PSNR&SSIM&Time\\
					\toprule			
					PatchDM&\XSolidBrush&$\X_t$&$\X_d$&32.08&0.937&61.27s\\
					WaveDM$_1$&\XSolidBrush&$\x_t$&$\x_d$&17.16&0.391&1.12s\\ 				
					WaveDM$_2$&\XSolidBrush&$\x_t^l$&$\x_d^l$&29.80&0.924&\textbf{0.75s}\\ 
					WaveDM$_3$&\CheckmarkBold&$\x_t^l$&$\x_d,\tilde{\x}_0^h$&\textbf{32.23}&\textbf{0.944}&0.97s\\ 
					\bottomrule 
			\end{tabular}}
		\end{center}
		
	\end{table}

	\begin{table}[ht]
		\small
		\renewcommand\arraystretch{1.1}
		\begin{center}
			\caption{Model configurations and parameter choices.}
			\resizebox{0.95\linewidth}{!}{
				\begin{tabular}{c|cc}
					\toprule
					Network&Setting&Time\\
					\toprule		
					\multirow{5}*{\makecell[c]{Noise \\Estimation\\ Network}}&Base channels&128\\
					~&Channel multipliers&$\{$1, 1, 2, 2, 4, 4$\}$\\
					~&Residual blocks per resolution &2\\
					~&Attention resolutions&$h/4$ ($h$: input height)\\
					~&Time step embedding length &512\\
					\midrule
					\multirow{4}*{HFRM}&Base channels&32\\
					~&Channel multipliers&$\{$1, 2, 4, 8, 16$\}$\\
					~&Residual blocks per resolution &1\\
					~&Attention resolutions&$h/4$ ($h$: input height)\\
					\bottomrule 
				\end{tabular}
			}
			\label{table_model_arch}
		\end{center}
		\vspace{0.6cm}
		
		\centering
		\begin{center}
			\renewcommand\arraystretch{1.1}
			\caption{Evaluation of two modules in terms of PSNR on the Raindrop dataset.}
			\resizebox{0.98\linewidth}{!}{
				\begin{tabular}{c|c|c|c}
					\toprule
					NEN & HFRM & Description & PSNR ($\uparrow$) \\
					\midrule
					\multirow{2}*{Default} & \multirow{2}*{Default}& Base channels (NEN): 128, Multipliers (NEN): $\{1, 1, 2, 2, 4, 4\}$ & \multirow{2}*{32.25dB} \\
					~&         ~& Base channels (HFRM): 32, Multipliers (HFRM): $\{1, 2, 4, 8, 16\}$ ~& \\
					\midrule
					\multirow{2}*{Variant 1} & \multirow{2}*{Default} & Base channels (NEN): 128, Multipliers (NEN): $\{1, 1, 2, 2, 4, 6\}$ & \multirow{2}*{32.37dB} \\
					~&         ~& Base channels (HFRM): 32, Multipliers (HFRM): $\{1, 2, 4, 8, 16\}$ ~& \\
					\midrule
					\multirow{2}*{Variant 2} & \multirow{2}*{Default} & Base channels (NEN): 256, Multipliers (NEN): $\{1, 1,  2, 2, 4, 6\}$ & \multirow{2}*{32.39dB} \\
					~&         ~& Base channels (HFRM): 32, Multipliers (HFRM): $\{1, 2, 4, 8, 16\}$ ~& \\
					\midrule
					\multirow{2}*{Default} & \multirow{2}*{Variant 1} & Base channels (NEN): 128, Multipliers (NEN): $\{1, 1, 2, 2, 4, 4\}$ & \multirow{2}*{32.22dB} \\
					~&           ~& Base channels (HFRM): 32 ~& \\
					\bottomrule
				\end{tabular}
			}
		\end{center}
		\label{table_model_evaluation}
	\end{table}

	\begin{figure}[t]
		\setlength{\abovecaptionskip}{0.1cm}
		\centering
		\includegraphics[width=0.472\textwidth]{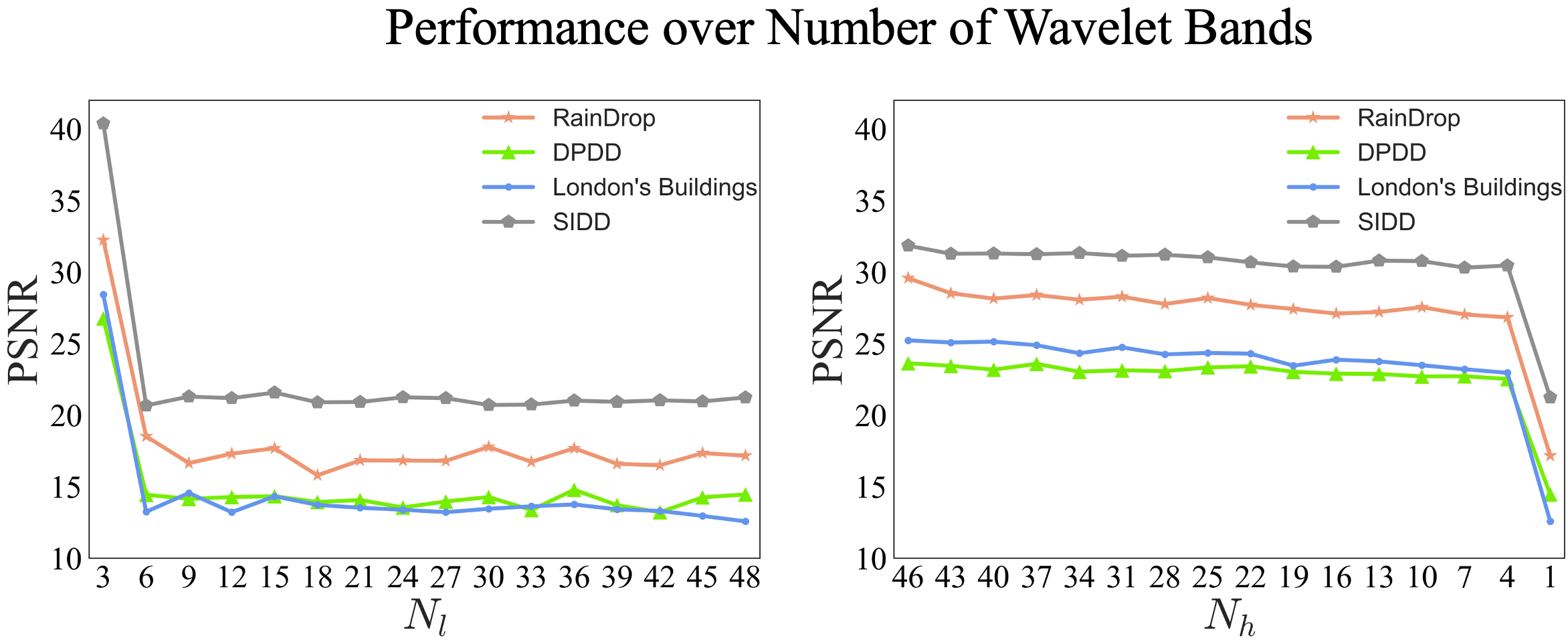}
		\caption{Different numbers of the wavelet bands for diffusion. $N_l$ indicates using the 1st to the $N_l$-th bands.	$N_h$ indicates using the 48-th to the $N_h$-th bands.}
		\label{bandchoice}
	\end{figure}

	\begin{figure}[ht]
		\small
		\centering
		\includegraphics[width=0.49\textwidth]{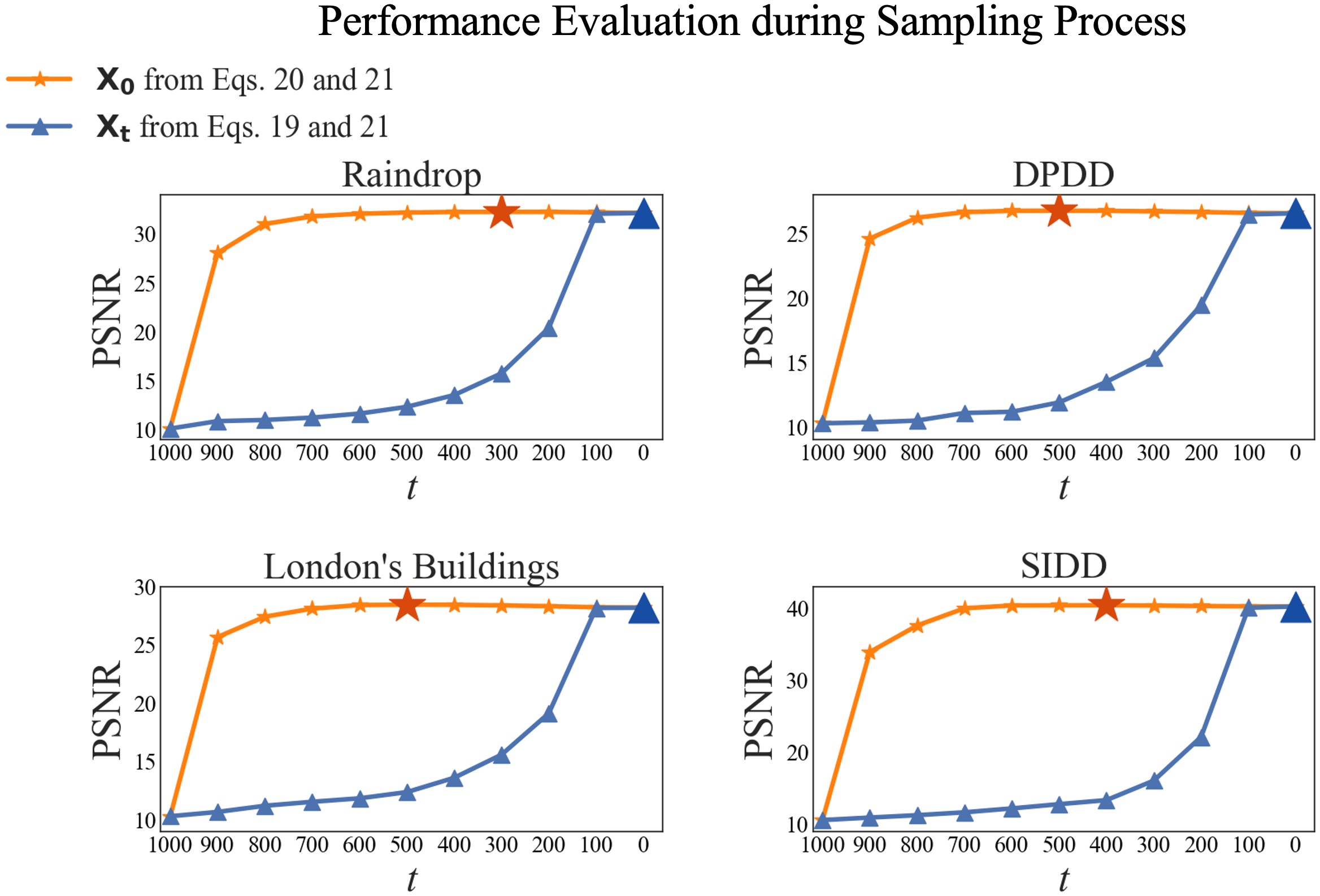}
		\caption{Performance evaluation during sampling process using 10 steps of stride 100 on four datasets.The \textcolor[RGB]{217,73,11}{\ding{72}} and \textcolor[RGB]{23, 78, 163}{\tiny{\TriangleUp}} denote the best PSNR values of the obtained $\X_0$ from Eq.~\ref{eq_ecs_wavedm} and \ref{eq_iwt} and $\X_t$ from Eq.~\ref{eq_ddim_wavedm} and \ref{eq_iwt}, respectively. $t$ represents the current time moment of sampling.}
		\label{sampling_overall}
	\end{figure}
	
	\begin{figure}[ht]
		\small
		\centering
		\includegraphics[width=0.49\textwidth]{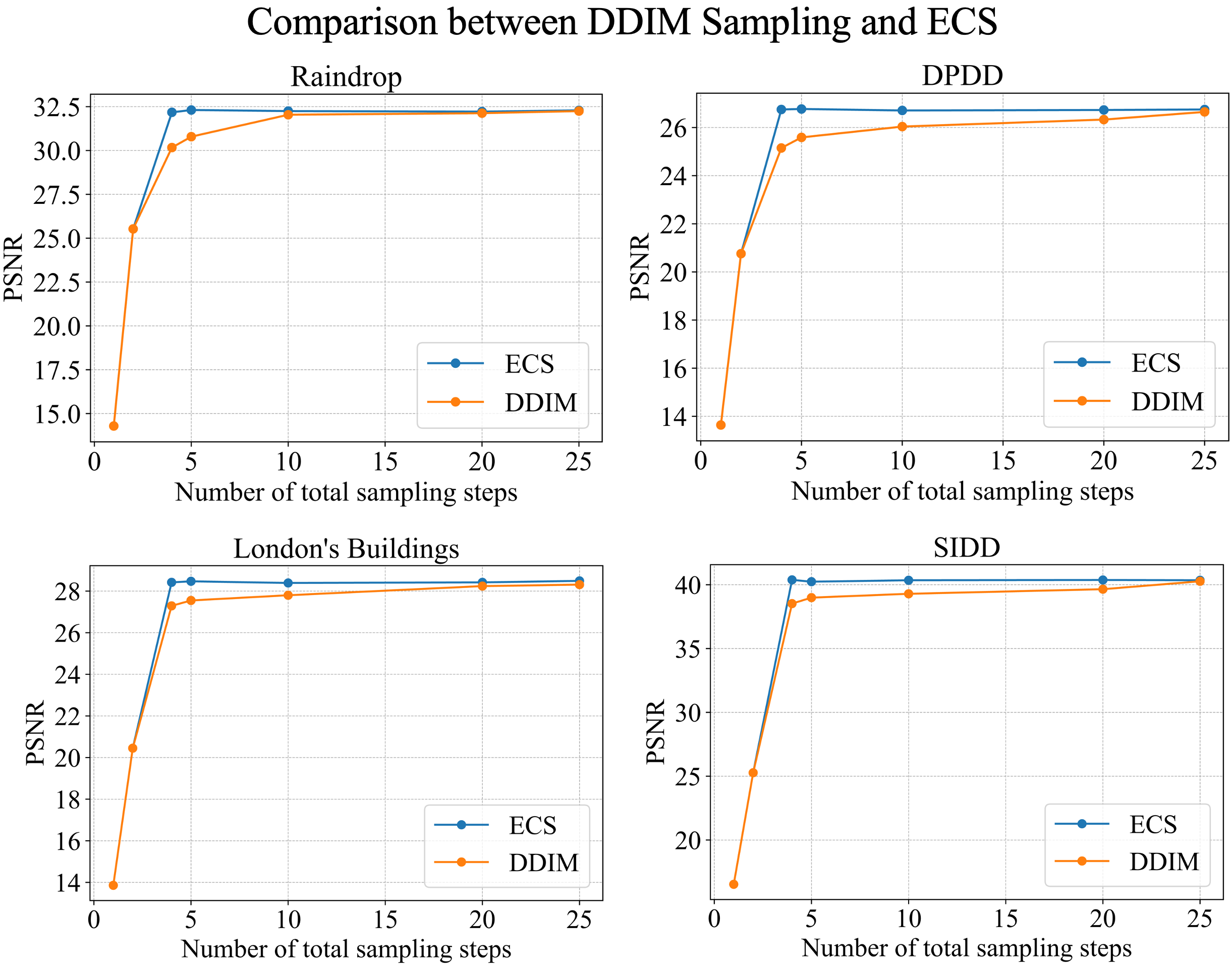}
		\caption{Restoration performance comparison between DDIM sampling and ECS under multiple sampling step settings on four datasets. $t$ represents the current time moment of sampling.}
		\label{fig_ecs_ddim}
	\end{figure}
	
	\begin{table}[h]
		\scriptsize
		\renewcommand\arraystretch{1.1}
		\begin{center}
			\caption{Performance comparison between DDIM sampling and ECS under two sampling step settings on four datasets. The PSNR values are computed at $t=0$ averaged on each dataset.}\label{table_sampling_trajectory}
			\resizebox{0.98\linewidth}{!}{
				\begin{tabular}{c|c|c|cccc}
					\toprule
					\multirow{2}*{Method}&\multirow{2}*{Step}&\multirow{2}*{Sampling Trajectory}&\multicolumn{4}{c}{PSNR}\\
					\cline{4-7}
					~&~&~&Raindrop&DPDD&London's Buildings&SIDD\\
					\toprule
					
					ECS&\multirow{4}*[4pt]{\makecell[c]{4}} &\makecell[c]{1000$\rightarrow$870$\rightarrow$730\\$\rightarrow$\textbf{600}\bm{$\rightarrow$}0}&32.19dB&26.75dB&28.42dB&40.38dB\\
					\cmidrule(l{2pt}r{2pt}){1-1}
					\cmidrule(l{2pt}r{2pt}){3-7}
					DDIM&~&\makecell[c]{1000$\rightarrow$750$\rightarrow$500\\$\rightarrow$250$\rightarrow$0}&30.17dB&25.15dB&26.95dB&38.52dB\\
					\midrule
					
					ECS&\multirow{4}*[4pt]{\makecell[c]{5}} &\makecell[c]{1000$\rightarrow$900$\rightarrow$800\\$\rightarrow$700$\rightarrow$\textbf{600}\bm{$\rightarrow$}0}&32.21dB&26.77dB&28.47dB&40.24dB\\
					\cmidrule(l{2pt}r{2pt}){1-1} \cmidrule(l{2pt}r{2pt}){3-7}
					DDIM&~&\makecell[c]{1000$\rightarrow$800$\rightarrow$600\\$\rightarrow$400$\rightarrow$200$\rightarrow$0}&30.79dB&25.59dB&27.29dB&38.99dB\\
					\bottomrule
				\end{tabular}
			}
			
		\end{center}
	\end{table}

	\begin{figure*}[t]
		\small
		\centering
		\begin{minipage}[t]{0.03\linewidth}
			\centering
			\raisebox{4.0\height}{$\X_t$}
		\end{minipage}
		\begin{minipage}[t]{0.1559\linewidth}
			\centerline{\includegraphics[width=2.88cm]{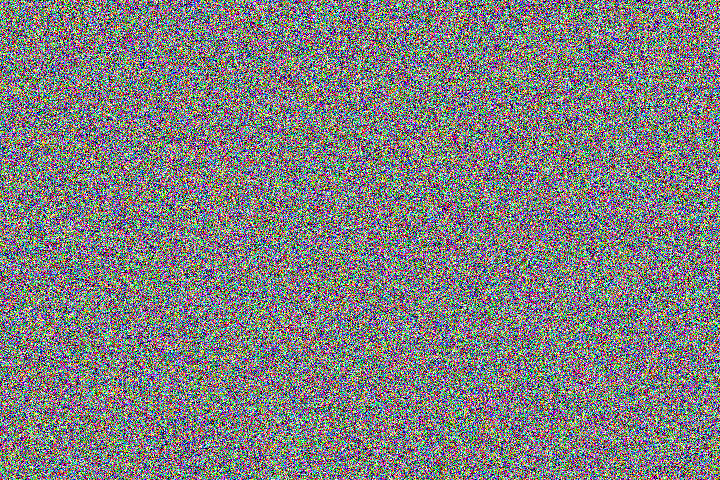}}
		\end{minipage}
		\begin{minipage}[t]{0.1559\linewidth}
			\centerline{\includegraphics[width=2.88cm]{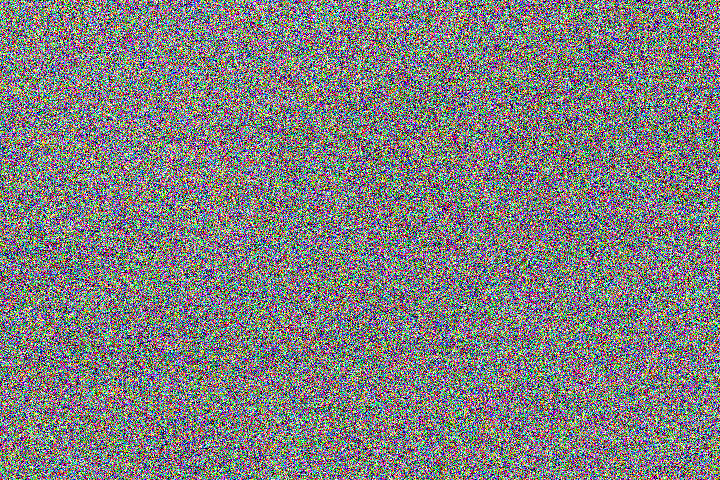}}
		\end{minipage}
		\begin{minipage}[t]{0.1559\linewidth}
			\centerline{\includegraphics[width=2.88cm]{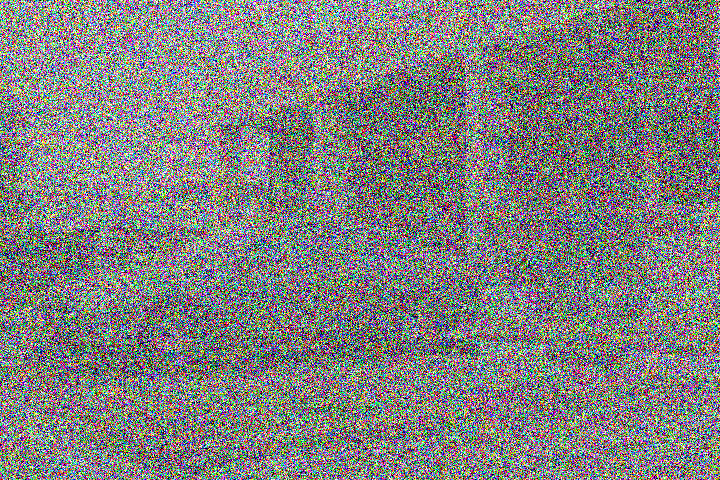}}
		\end{minipage}
		\begin{minipage}[t]{0.1559\linewidth}
			\centerline{\includegraphics[width=2.88cm]{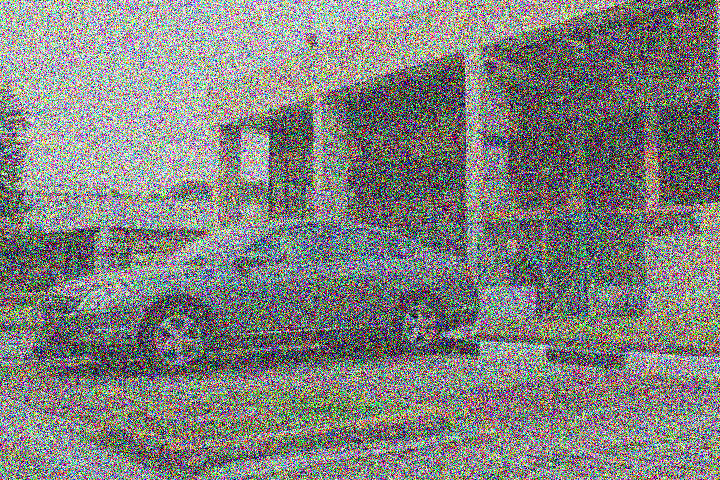}}
		\end{minipage}
		\begin{minipage}[t]{0.1559\linewidth}
			\centerline{\includegraphics[width=2.88cm]{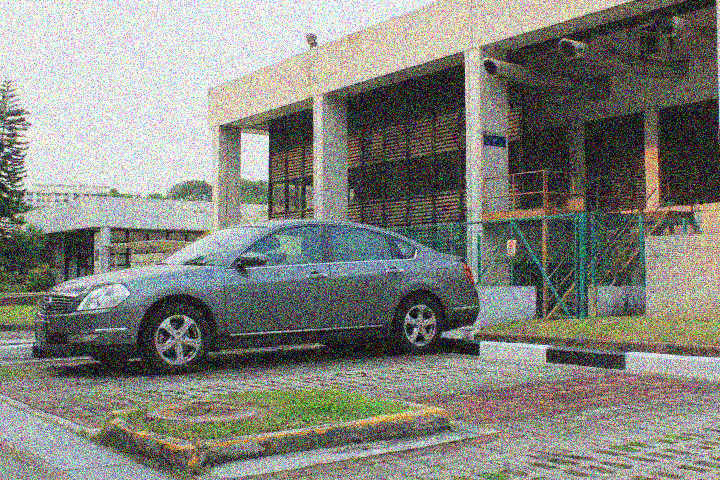}}
		\end{minipage}
		\begin{minipage}[t]{0.1559\linewidth}
			\centerline{\includegraphics[width=2.88cm]{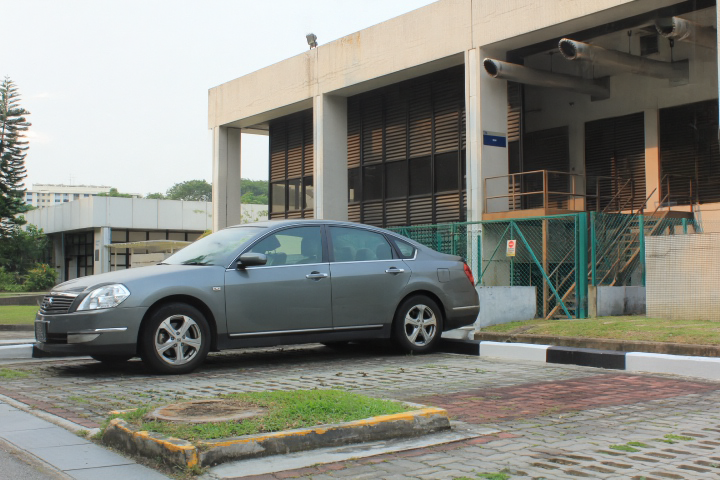}}
		\end{minipage}
		\hfill
		\begin{minipage}[t]{0.03\linewidth}
			\centering
			\raisebox{4.0\height}{$\X_0$}
			\centerline{$t$}
		\end{minipage}
		\begin{minipage}[t]{0.1559\linewidth}
			\centerline{\includegraphics[width=2.88cm]{ablations/ecs/visual/55_rain_output_x00.png}}
			\centerline{$1000$}
		\end{minipage}
		\begin{minipage}[t]{0.1559\linewidth}
			\centerline{\includegraphics[width=2.88cm]{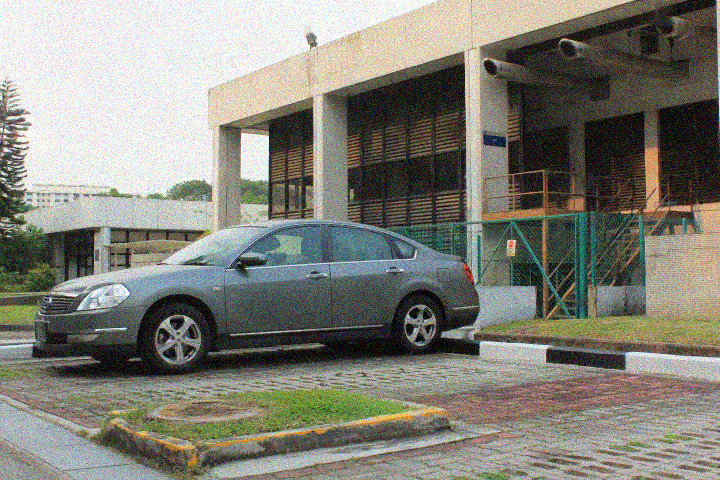}}
			\centerline{$800$}
		\end{minipage}
		\begin{minipage}[t]{0.1559\linewidth}
			\centerline{\includegraphics[width=2.88cm]{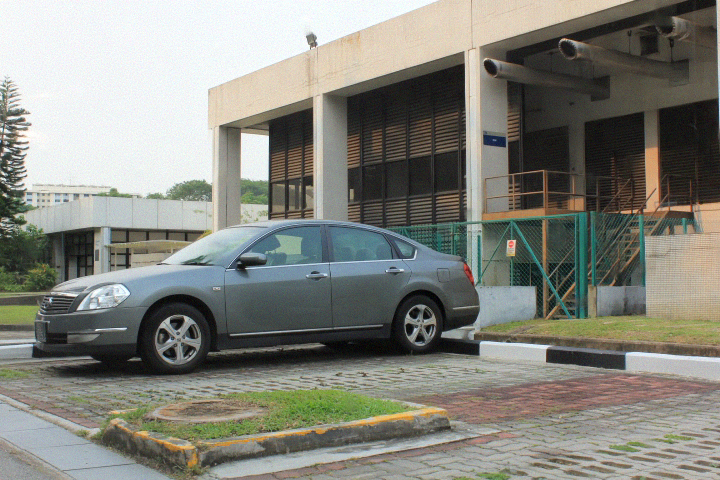}}
			\centerline{$600$}
		\end{minipage}
		\begin{minipage}[t]{0.1559\linewidth}
			\centerline{\includegraphics[width=2.88cm]{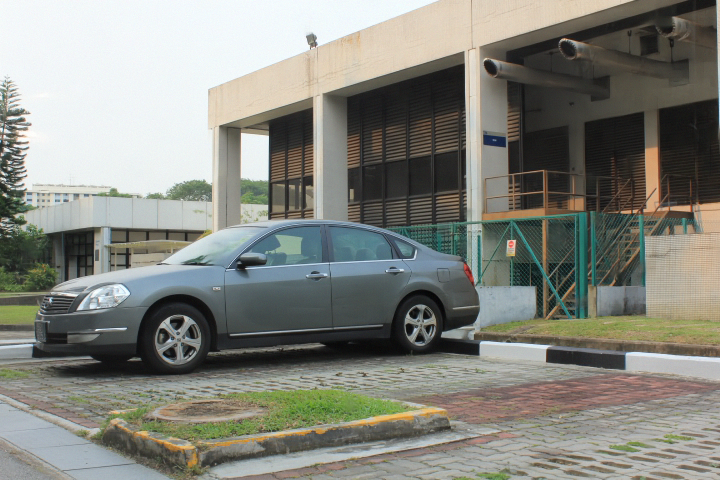}}
			\centerline{$400$}
		\end{minipage}
		\begin{minipage}[t]{0.1559\linewidth}
			\centerline{\includegraphics[width=2.88cm]{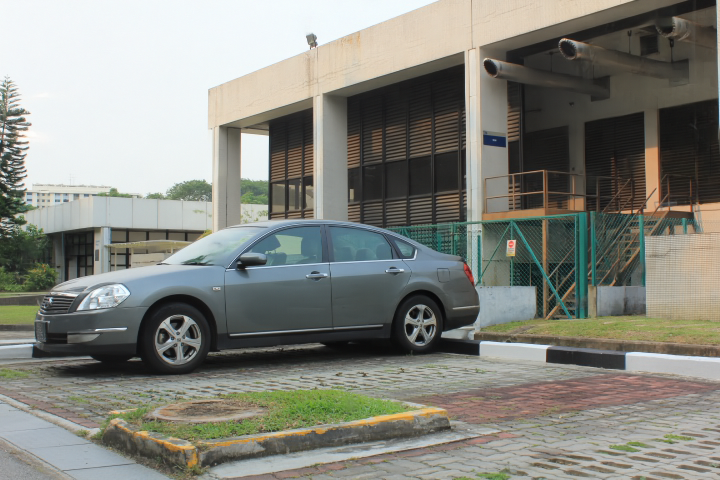}}
			\centerline{$200$}
		\end{minipage}
		\begin{minipage}[t]{0.1559\linewidth}
			\centerline{\includegraphics[width=2.88cm]{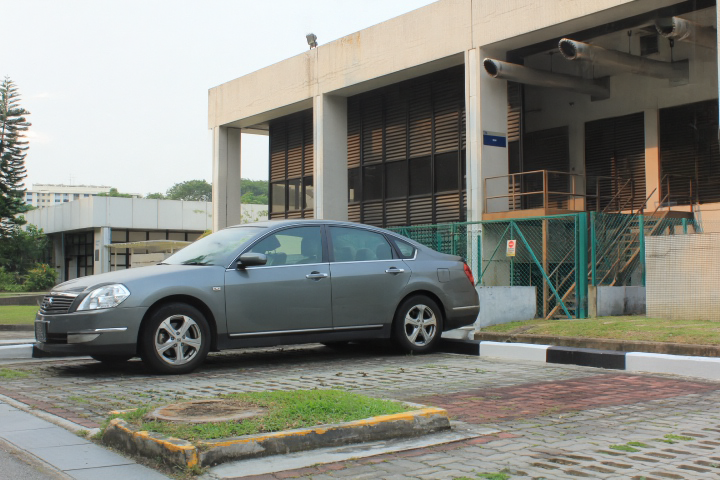}}
			\centerline{$0$}
		\end{minipage}
		\caption{Visual results of the $\X_t$ from Eq.~\ref{eq_ddim_wavedm} and \ref{eq_iwt}, and $\X_0$ from Eq.~\ref{eq_ecs_wavedm} and \ref{eq_iwt} during the sampling process for image raindrop removal.}
		\label{fig_ecs_ddim_visual}
		\vspace{0.2cm}
		\small
		\centering
		\begin{minipage}[t]{0.1199\linewidth}			\centerline{\includegraphics[width=2.23cm]{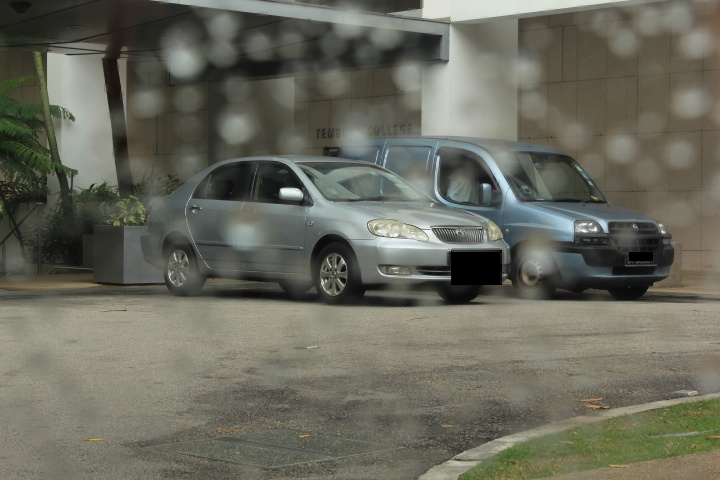}}
			\centerline{Degraded}
		\end{minipage}
		\begin{minipage}[t]{0.1199\linewidth}			\centerline{\includegraphics[width=2.23cm]{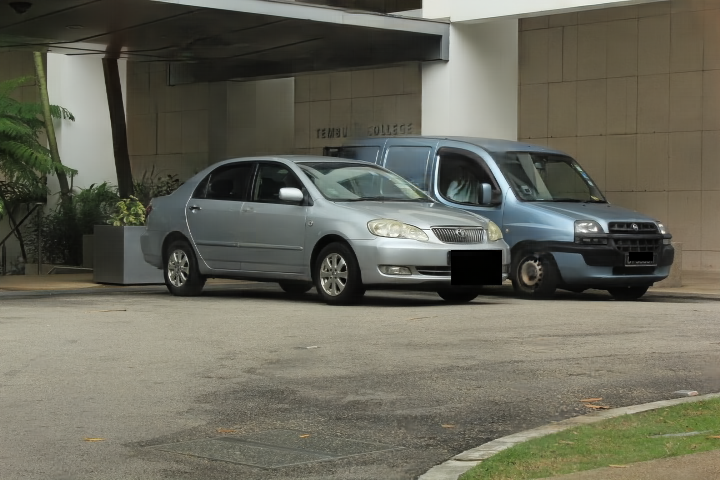}}
			\centerline{30.45dB}
		\end{minipage}
		\begin{minipage}[t]{0.1199\linewidth}			\centerline{\includegraphics[width=2.23cm]{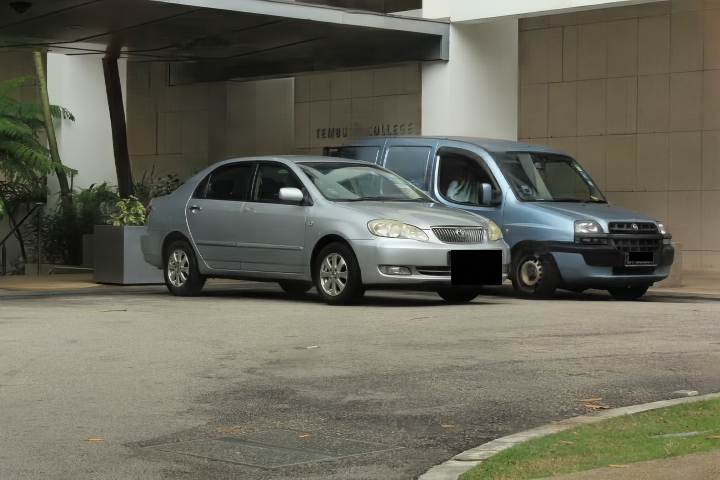}}
			\centerline{30.24dB}
		\end{minipage}
		\begin{minipage}[t]{0.1199\linewidth}			\centerline{\includegraphics[width=2.23cm]{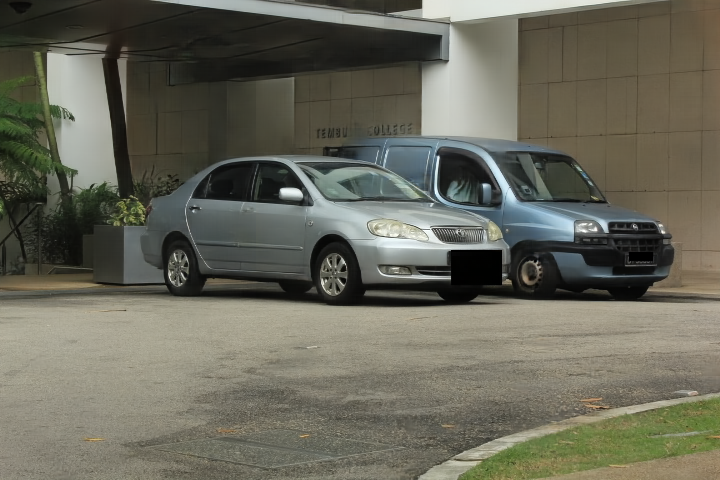}}
			\centerline{30.34dB}
		\end{minipage}
		\begin{minipage}[t]{0.1199\linewidth}			\centerline{\includegraphics[width=2.23cm]{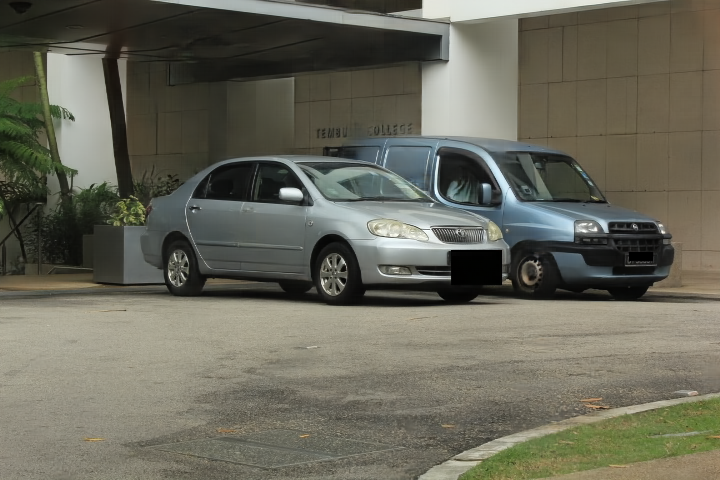}}
			\centerline{30.37dB}
		\end{minipage}
		\begin{minipage}[t]{0.1199\linewidth}			\centerline{\includegraphics[width=2.23cm]{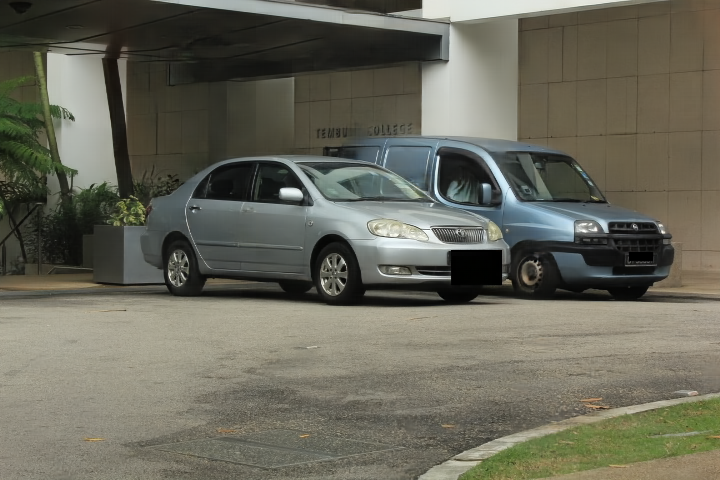}}
			\centerline{30.34dB}
		\end{minipage}
		\begin{minipage}[t]{0.1199\linewidth}			\centerline{\includegraphics[width=2.23cm]{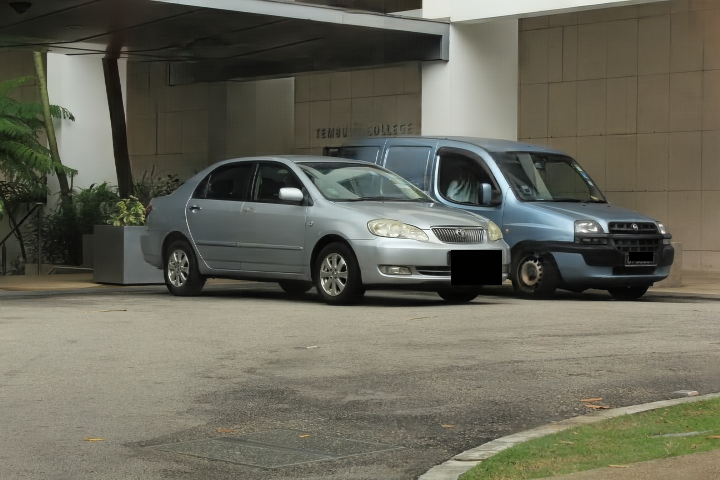}}
			\centerline{30.43dB}
		\end{minipage}
		\begin{minipage}[t]{0.1199\linewidth}			\centerline{\includegraphics[width=2.23cm]{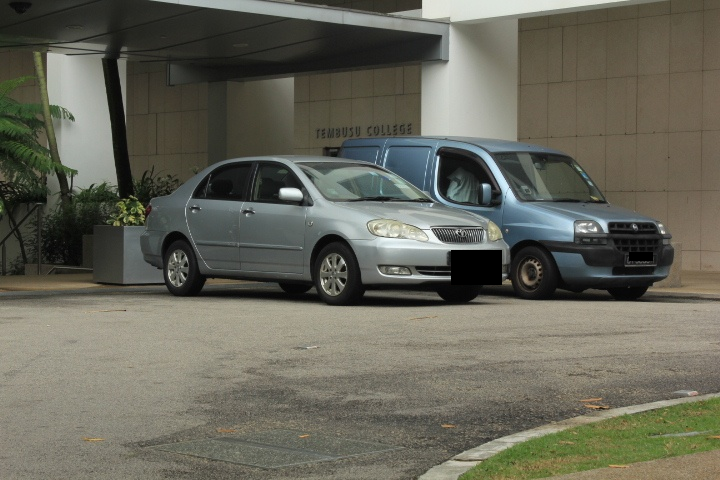}}
			\centerline{PSNR}       
		\end{minipage}
		\hfill
		\begin{minipage}[t]{0.1199\linewidth}			\centerline{\includegraphics[width=2.23cm]{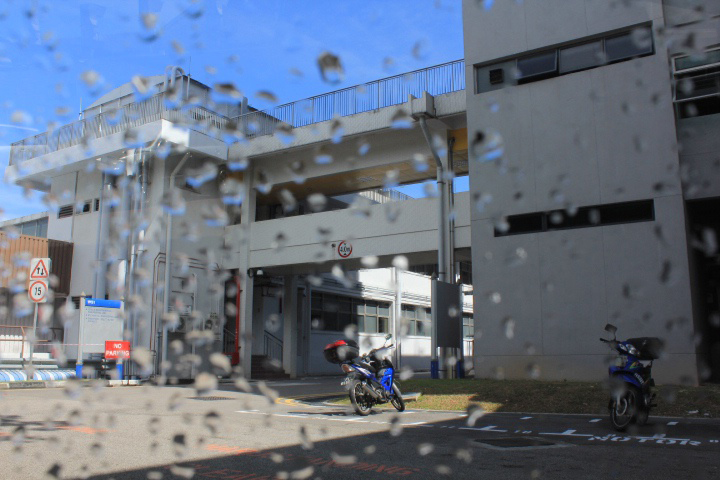}}
			\centerline{Degraded}
			\centerline{Seed number}
		\end{minipage}
		\begin{minipage}[t]{0.1199\linewidth}			\centerline{\includegraphics[width=2.23cm]{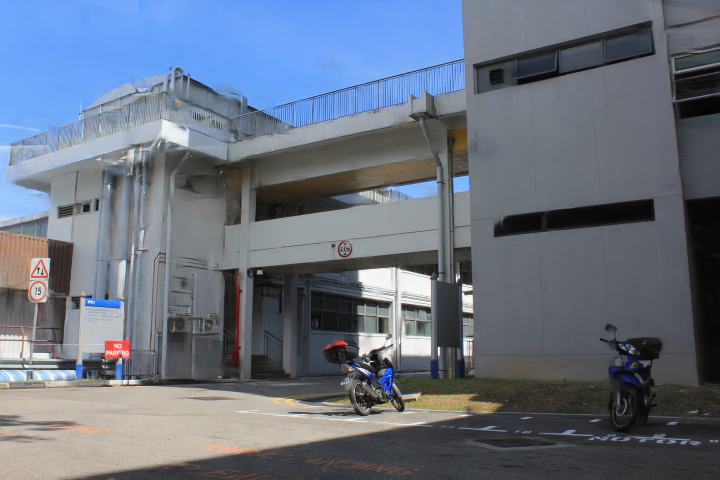}}
			\centerline{27.02dB}
			\centerline{23255}
		\end{minipage}
		\begin{minipage}[t]{0.1199\linewidth}			\centerline{\includegraphics[width=2.23cm]{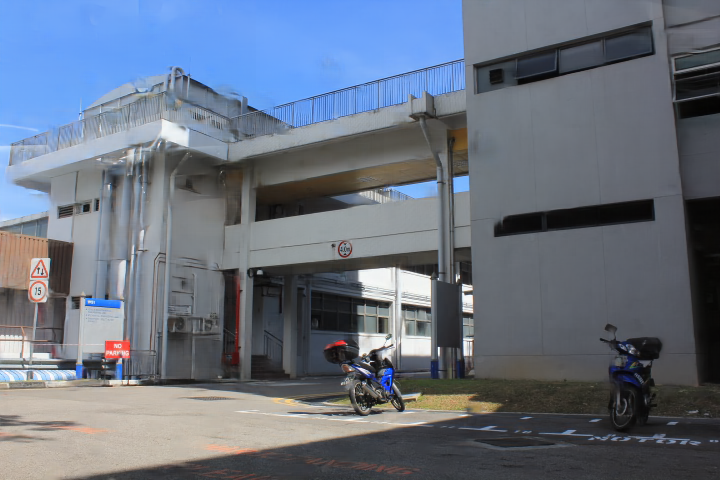}}
			\centerline{27.04dB}
			\centerline{64442}
		\end{minipage}
		\begin{minipage}[t]{0.1199\linewidth}			\centerline{\includegraphics[width=2.23cm]{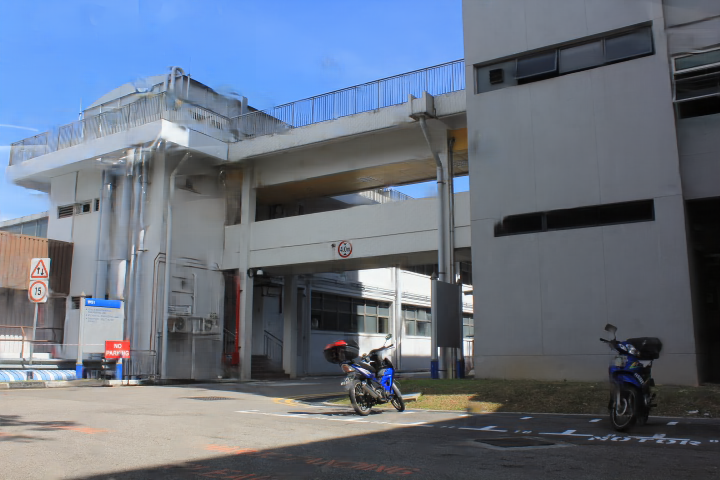}}
			\centerline{27.11dB}
			\centerline{3424}
		\end{minipage}
		\begin{minipage}[t]{0.1199\linewidth}			\centerline{\includegraphics[width=2.23cm]{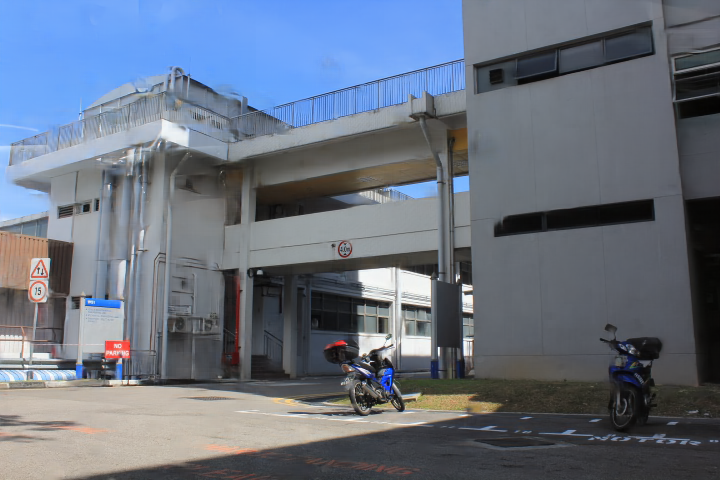}}
			\centerline{27.08dB}
			\centerline{6179}
		\end{minipage}
		\begin{minipage}[t]{0.1199\linewidth}			\centerline{\includegraphics[width=2.23cm]{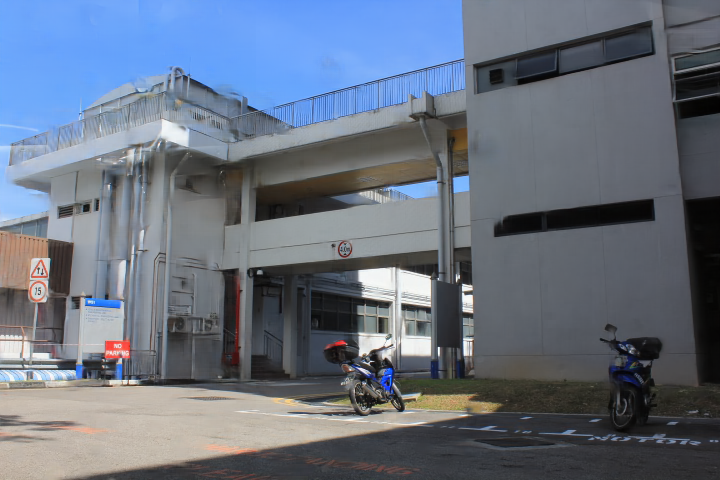}}
			\centerline{27.09dB}
			\centerline{33456}
		\end{minipage}
		\begin{minipage}[t]{0.1199\linewidth}			\centerline{\includegraphics[width=2.23cm]{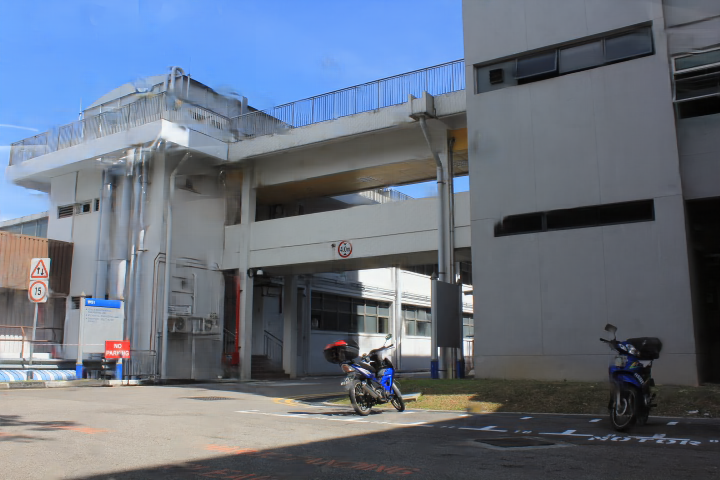}}
			\centerline{27.08dB}
			\centerline{34524}
		\end{minipage}
		\begin{minipage}[t]{0.1199\linewidth}			\centerline{\includegraphics[width=2.23cm]{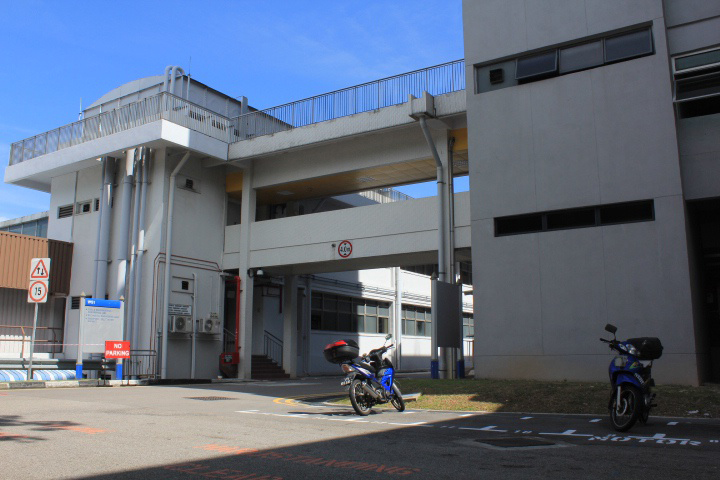}}
			\centerline{PSNR}       
			\centerline{GT}
		\end{minipage}
		\caption{Visual results of the generated samples with different seeds for image raindrop removal. Each column is generated from the same random seed.}
		\label{fig_seed}
	\end{figure*}
	
	\begin{figure}[ht]

		\centering
		\includegraphics[width=0.42\textwidth]{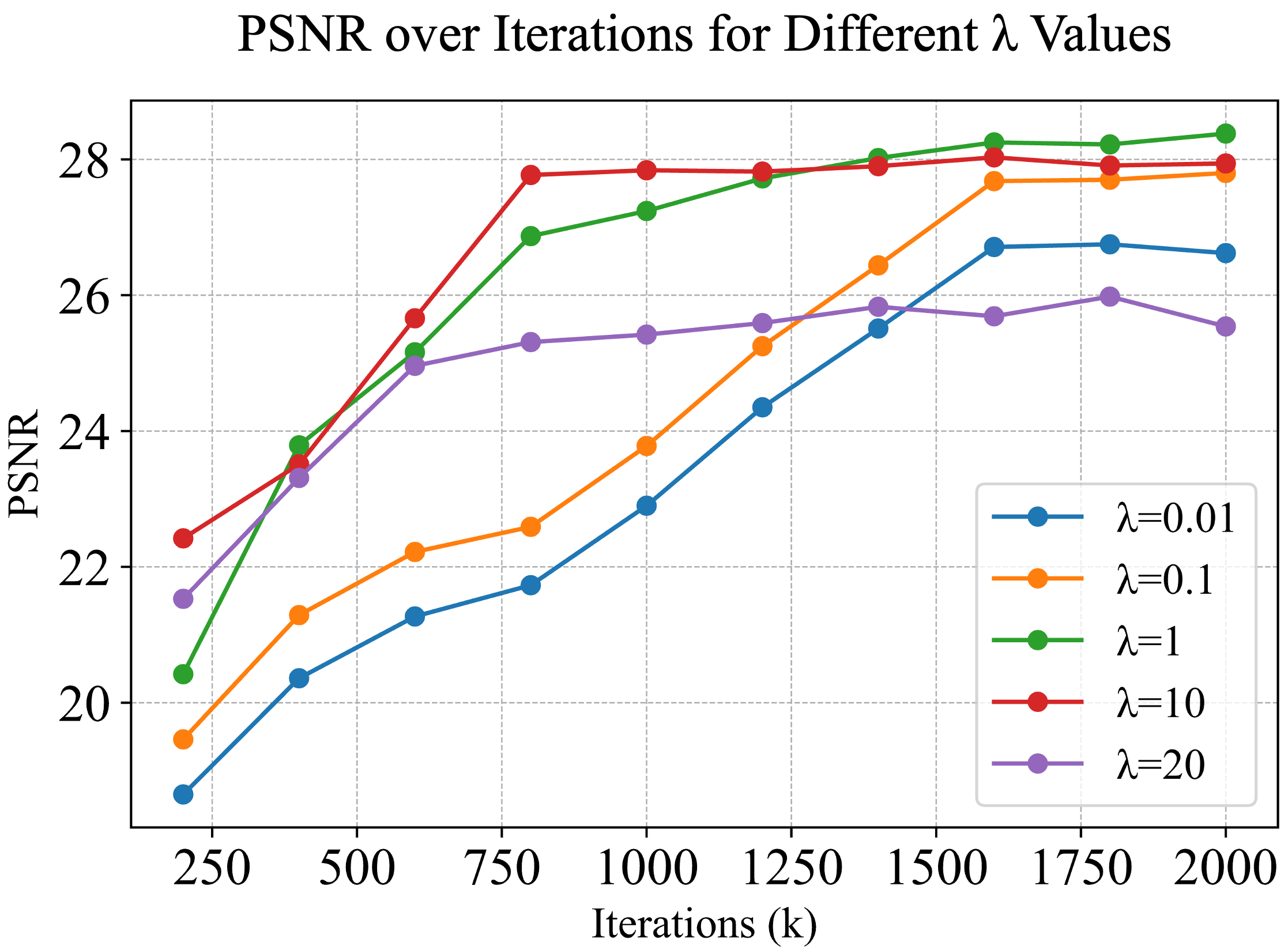}
		\caption{Performance comparison in terms of PSNR across different training iterations for varying values of \(\lambda\) on the London's Buildings dataset.}
		\label{fig_lambda}
	\end{figure}

	\subsection{Ablation Studies}

	\subsubsection{Input Conditions}
	In this section, we explore how the varieties of conditions influence the restoration performance. Several choices for them with corresponding quantitative results are shown in Table~\ref{table1}, in which all methods use 25 steps of the DDIM sampling. According to them, we can see that although PatchDM performs well in the spatial domain, the processing of a large number of small-size patches is extremely time-consuming. Instead, when switching to the wavelet domain, the total sampling time is reduced by around $1/60$ due to the small spatial size after wavelet transform. However, when modeling the distribution on all 48 bands ($\x_t$) of clean images with all-frequency components ($\x_d$) of degraded images as the condition (WaveDM$_1$), the results are the worst.
	WaveDM$_2$ models the distribution of clean images on the first three low-frequency bands ($\x_t^l$) with the corresponding bands ($\x_d^l$) of degraded images as the condition, of which the performance 
	is much better than WaveDM$_1$. WaveDM$_3$ is our full WaveDM, where an additional HFRM is added to estimate the high-frequency bands ($\tilde{\x}_0^h$) of clean images, which serves as not only the essential high-frequency bands for inverse wavelet transform but also an extra condition to predict the low-frequency bands. The PSNR/SSIM gain by WaveDM$_3$ verifies that HFRM is effective with little extra computation (one-pass for $\tilde{\x}_0^h$).

	\subsubsection{Model Configurations}
	In this part, we explore the significance of structures of the Noise Estimation Network (NEN) and HFRM within our WaveDM, particularly focusing on restoration performance. 
	The default configurations for these modules are detailed in Table~\ref{table_model_arch}. Both modules have the U-Net architecture similar to PatchDM \cite{ozdenizci2022restoring}. Specifically, the setting of NEN is the same as PatchDM’s. Meanwhile, HFRM uses fewer blocks and a reduced number of feature channels for a lightweight design. To further understand the influence of these modules, we devise two alternative configurations of NEN and one for HFRM. These variants are then evaluated on the Raindrop dataset. The comparative results, presented in Table \ref{table_model_evaluation},  indicate that the performance slightly drops when NEN scales down (compare rows 1--3) and the complexity of HFRM has little impact on the restoration quality (compare rows 1 and 4).

	\subsubsection{Wavelet Bands}
	To further explore what wavelet bands should be used in the diffusion model, we select $N$ wavelet bands for diffusion, and the other $48-N$ bands in $\x_d$ serve as the input to HFRM. Experimental comparison on the four datasets is shown in Fig.~\ref{bandchoice}, from which we can observe that the restoration performance reaches the best when modeling the first three low-frequency bands for diffusion. Therefore, this setting is used in all the following experiments.\label{bandchoiceanalysis}

	\subsubsection{Wavelet Levels}
	We also conduct an experiment to explore the effect of different wavelet transform levels on the restoration performance. For levels 1, 2 and 3, the values of PSNR/time of WaveDM with the 25-step DDIM sampling on London's Buildings are  28.12dB/126.16s, 28.39dB/5.21s and 24.14dB/0.72s. After the 1-level Haar wavelet transform, the wavelet bands still have a large spatial size and also need to be cut into patches for processing, which is time-consuming. However, in the 2-level wavelet transform, the inference time can be reduced to 5.21s with a similar PSNR. When the level is further increased (3 or higher), the performance is harmed because too many details are lost in the low-frequency bands for diffusion.

	\subsubsection{Efficient Conditional Sampling}
	\label{exp_ecs}

	The development of the ECS is based on a series of exploratory experiments. Fig.~\ref{sampling_overall}  provides a visual representation of a subset of these experiments. The finding from these experiments, serves as a foundation of the ECS formulation.
	Upon examining the results, as depicted in Fig.~\ref{sampling_overall}, we observe that when the sampling time reaches the moment around \(t=600\), the PSNR between the predicted $\X_0$ from $\X_{600}$ and the ground truth (GT) reaches a significant value, even if not the highest. After this moment ($t<600$), the PSNR remains relatively stable with negligible fluctuations. Based on these experiments, we select \(M=600\) as the default setting for ECS for all experiments. 
	Besides, the PSNR of predicted $\X_0$ decreases when the sampling continues after \textcolor[RGB]{217,73,11}{\ding{72}}, the reason of which comes from the fact that the second term $\sqrt{1-\bar{\alpha}_{t-1}}\cdot\bm{\epsilon}_{\theta}(\x_t^l,\x_d,\tilde{\x}_0^h,t)$ in Eq.~\ref{eq_ecs_wavedm} introduces extra noise to the results. Due to the small value of the weight term $\sqrt{1-\bar{\alpha}_{t-1}}$, the decline in PSNR is slight. We also present an example of the denoising process for synthesizing clean images in Fig.~\ref{fig_ecs_ddim_visual}.
	To compare ECS with the conventional DDIM sampling method, we show PSNR values for both strategies with the same number of total sampling steps but with different trajectories. Table~\ref{table_sampling_trajectory} highlights this comparison for two exemplary settings. An extended comparison with more settings of sampling steps is presented in Fig.~\ref{fig_ecs_ddim}. From the results, it is observable that as the sampling steps increase, the difference in PSNR  decreases. However, when the number of total sampling steps is set below 10, ECS shows better
	performance, demonstrating its superior efficiency over the traditional DDIM sampling.
	
	\begin{table*}[ht]
		\renewcommand\arraystretch{1.025}
		\begin{center}
			\caption{Quantitative comparison with SOTA methods on various restoration tasks.}
			\vspace{-0.1cm}
			\label{table_sota}
			\setlength{\tabcolsep}{3.0mm}{
				\begin{tabular}{c|c|c|c|cccccc}
					
					\toprule
					\small
					Task&Type&Method&Step&PSNR$\uparrow$&SSIM$\uparrow$&FID$\downarrow$&Time$\downarrow$&Parameters$\downarrow$&Memory$\downarrow$\\
					\toprule

					\multirow{7}*{\makecell[c]{Raindrop \\ Removal}}&\multirow{5}*{One-pass}&DuRN\cite{liu2019dual}&\multirow{5}*{1}&31.24dB&0.926&30.63&\textbf{0.09s}&10.2M&4108MB\\

					~&~&CCN\cite{quan2021removing}&~&31.44dB&0.947&28.94&0.80s&12.4M&\underline{3738MB}\\
					~&~&RainAttn\cite{quan2019deep}&~&31.44dB&0.926&28.47&0.41s&\textbf{6.24M}&8335MB\\
					~&~&AttnGAN\cite{qian2018attentive}&~&31.59dB&0.917&27.84&0.63s&\underline{7.08M}&8797MB\\				
					~&~&IDT\cite{xiao2022image}&~&31.87dB&0.931&25.51&0.39s&16.4M&\textbf{3660MB}\\
					\cmidrule{2-10}

					~&\multirow{3}*{Iterative}&PatchDM$_{64}$\cite{ozdenizci2022restoring}&10&32.13dB&0.939&25.12&24.36s&110M&8758MB\\
					~&~&PatchDM$_{128}$\cite{ozdenizci2022restoring}&50&\textbf{32.31dB}&\underline{0.946}&\textbf{20.57}&301.35s&110M&22313MB\\
					~&~&Ours&8&\underline{32.25dB}&\textbf{0.948}&\underline{23.53}&\underline{0.30s}&124M&4965MB\\

					\bottomrule
					\specialrule{0em}{0.6pt}{0.6pt}

					\toprule

					\multirow{7}*{\makecell[c]{Rain steaks \\ Removal}}&\multirow{5}*{One-pass}&HRGAN\cite{li2019heavy}&\multirow{5}*{1}&21.56dB&0.855&69.25&0.80s&50.4M&\underline{3663MB}\\

					~&~&PCNet\cite{jiang2021pcnet}&~&26.19dB&0.901&44.57&\textbf{0.06s}&\textbf{0.63M}&\textbf{1845MB}\\				
					~&~&MPRNet\cite{zamir2021multi}&~&28.03dB&0.919&30.61&\underline{0.12s}&\underline{20.1M}&6942MB\\
					~&~&All-in-One\cite{li2020all}&~&24.71dB&0.898&$\backslash$&$\backslash$&$\backslash$&$\backslash$\\				
					~&~&TransWeather\cite{valanarasu2022transweather}&~&\underline{28.83dB}&0.900&22.52&0.16s&38.1M&4734MB\\
					\cmidrule{2-10}

					~&\multirow{2}*{Iterative}&PatchDM$_{64}$\cite{ozdenizci2022restoring}&25&28.38dB&\underline{0.932}&\underline{17.36}&59.97s&110M&8759MB\\

					~&~&Ours&4&\textbf{31.39dB}&\textbf{0.943}&\textbf{11.42}&0.16s&124M&4965MB\\

					\bottomrule
					\specialrule{0em}{0.6pt}{0.6pt}

					\toprule

					\multirow{7}*{Dehazing}&\multirow{5}*{One-pass}&DCP\cite{he2010single}&\multirow{5}*{1}&19.13dB&0.815&20.03&\textbf{0.05s}&$\backslash$&2388MB\\

					~&~&GridDehazeNet\cite{liu2019griddehazenet}&~&30.86dB&0.982&4.76&0.20s&\textbf{0.96M}&1956MB\\
					~&~&MSBDN\cite{dong2020multi}&~&33.48dB&0.982&5.59&0.16s&31.4M&\textbf{1756MB}\\
					~&~&FFA-Net\cite{qin2020ffa}&~&33.57dB&0.984&6.43&0.36s&4.46M&2246MB\\				
					~&~&DehazeFormer-B\cite{song2023vision}&~&34.95dB&0.984&\underline{4.58}&\underline{0.14s}&\underline{2.51M}&\underline{1760MB}\\
					\cmidrule{2-10}

					~&\multirow{2}*{Iterative}&PatchDM$_{64}$\cite{ozdenizci2022restoring}&25&\underline{35.52dB}&\underline{0.989}&5.75&19.31s&110M&8759MB\\
					~&~&Ours&4&\textbf{37.00dB}&\textbf{0.994}&\textbf{2.80}&0.15s&124M&6336MB\\

					\bottomrule
					\specialrule{0em}{0.6pt}{0.6pt}

					\toprule
					\multirow{7}*{\makecell[c]{Single-pixel \\ Defocus\\Debluring}}&\multirow{5}*{One-pass}&DMENet\cite{lee2019deep}&\multirow{5}*{1}&23.41dB&0.714&54.51&1.79s&26.9M&\underline{8954MB}\\
					~&~&DPDNet\cite{abuolaim2020defocus}&~&24.34dB&0.747&55.21&\underline{0.32s}&32.3M&11747MB\\
					~&~&KPAC\cite{son2021single}&~&25.22dB&0.774&46.49&0.33s&\textbf{2.64M}&12575MB\\
					~&~&IFAN\cite{lee2021iterative}&~&25.37dB&0.789&46.47&\textbf{0.20s}&10.5M&19273MB\\
					~&~&Restormer\cite{zamir2022restormer}&~&25.98dB&0.811&\textbf{43.13}&3.22s&\underline{25.5M}&26256MB\\
					\cmidrule{2-10}

					~&\multirow{2}*{Iterative}&PatchDM$_{64}$\cite{ozdenizci2022restoring}&25&\underline{26.49dB}&\underline{0.812}&47.92&365.20s&110M&\textbf{8759MB}\\

					~&~&Ours&4&\textbf{26.75dB}&\textbf{0.822}&\underline{45.43}&0.47s&124M&12190MB\\

					\bottomrule
					\specialrule{0em}{0.6pt}{0.6pt}

					\toprule
					%

					\multirow{5}*{\makecell[c]{Dual-pixel \\ Defocus\\Debluring}}&\multirow{5}*{One-pass}&DPDNet\cite{abuolaim2020defocus}&\multirow{5}*{1}&25.13dB&0.786&45.52&0.32s&32.3M&\textbf{12265MB}\\
					~&~&RDPD\cite{abuolaim2021learning}&~&25.39dB&0.772&39.71&\underline{0.29s}&\underline{24.3M}&18492MB\\
					~&~&IFAN\cite{lee2021iterative}&~&25.99dB&0.804&36.87&\textbf{0.20s}&\textbf{10.5M}&20127MB\\
					~&~&Restormer\cite{zamir2022restormer}&~&\underline{26.66dB}&\underline{0.833}&\underline{34.49}&3.22s&25.5M&28214MB\\
					\cmidrule{2-10}
					
					~&Iterative&Ours&4&\textbf{27.49dB}&\textbf{0.855}&\textbf{31.28}&0.48s&124M&\underline{12326MB}\\
					\bottomrule
					\specialrule{0em}{0.6pt}{0.6pt}
					\toprule
					
					\multirow{6}*{Demoiréing}&\multirow{4}*{One-pass}&MultiscaleNet\cite{sun2018moire}&\multirow{3}*{1}&23.64dB&0.791&71.39&0.59s&\textbf{0.65M}&15486MB\\
					~&~&WDNet\cite{liu2020wavelet}&~&24.12dB&0.847&51.65&\textbf{0.18s}&\underline{3.92M}&21472MB\\
					~&~&FHDe$^2$Net\cite{he2020fhde}&~&24.31dB&0.799&41.38&2.03s&13.6M&27686MB\\
					~&~&ESDNet\cite{yu2022towards}&~&25.67dB&0.871&58.92&\underline{0.23s}&5.93M&28432MB\\
					\cmidrule{2-10}
					
					~&\multirow{2}*{Iterative}&PatchDM$_{64}$\cite{ozdenizci2022restoring}&25&\underline{28.09dB}&\underline{0.934}&\underline{33.51}&656.75s&110M&\textbf{8759MB}\\
					~&~&Ours&4&\textbf{28.42dB}&\textbf{0.942}&\textbf{23.14}&1.01s&124M&\underline{13052MB}\\

					\bottomrule
					\specialrule{0em}{0.6pt}{0.6pt}

					\toprule

					\multirow{6}*{\makecell[c]{Real \\ Denoising}}&\multirow{4}*{One-pass}&MIRNet\cite{zamir2020learning}&\multirow{3}*{1}&39.72dB&0.959&47.71&0.090s&31.8M&5805MB\\
					~&~&MPRNet\cite{zamir2021multi}&~&39.71dB&0.958&49.54&\underline{0.055s}&\textbf{20.1M}&\textbf{2861MB}\\
					~&~&Uformer\cite{wang2022uformer}&~&39.77dB&0.959&\underline{47.17}&\textbf{0.031s}&50.9M&9157MB\\
					~&~&Restormer\cite{zamir2022restormer}&~&\underline{40.02dB}&\underline{0.960}&47.28&0.114s&\underline{25.5M}&7702MB\\
					\cmidrule{2-10}

					~&\multirow{2}*{Iterative}&PatchDM$_{64}$\cite{ozdenizci2022restoring}&25&39.86dB&0.959&47.59&9.332s&110M&8759MB\\
					~&~&Ours&4&\textbf{40.38dB}&\textbf{0.962}&\textbf{47.01}&0.062s&124M&\underline{3430MB}\\

					\bottomrule
				\end{tabular}
			}
			
		\end{center}
		\vspace{-0.15cm}
	\end{table*}
	
	\begin{figure*}[ht]
		\small
		\centering
		\begin{minipage}[t]{0.1199\linewidth}			\centerline{\includegraphics[width=2.23cm]{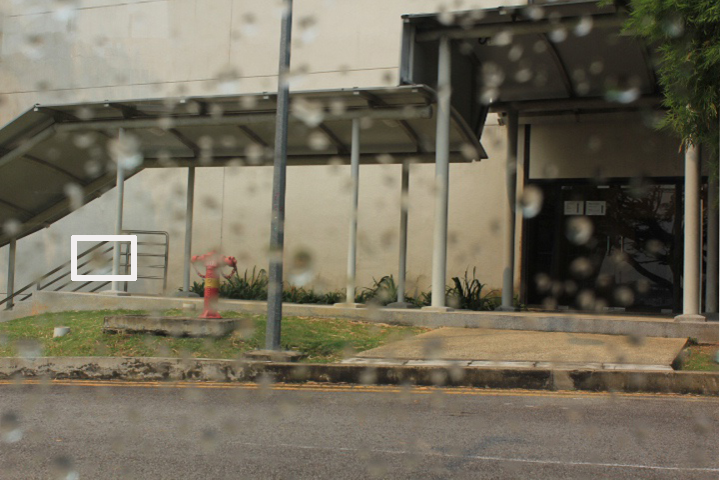}}
			\centerline{Degraded}
			\centerline{Image}
		\end{minipage}
		\begin{minipage}[t]{0.1199\linewidth}			\centerline{\includegraphics[width=2.23cm]{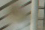}}
			\centerline{25.12dB}
			\centerline{Raindrop}
		\end{minipage}
		\begin{minipage}[t]{0.1199\linewidth}			\centerline{\includegraphics[width=2.23cm]{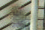}}
			\centerline{24.99dB}
			\centerline{AttnGAN\cite{qian2018attentive}}
		\end{minipage}
		\begin{minipage}[t]{0.1199\linewidth}			\centerline{\includegraphics[width=2.23cm]{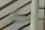}}
			\centerline{27.71dB}
			\centerline{DuRN\cite{liu2019dual}}
		\end{minipage}
		\begin{minipage}[t]{0.1199\linewidth}			\centerline{\includegraphics[width=2.23cm]{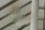}}
			\centerline{30.10dB}
			\centerline{RainAttn\cite{quan2019deep}}
		\end{minipage}
		\begin{minipage}[t]{0.1199\linewidth}			\centerline{\includegraphics[width=2.23cm]{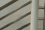}}
			\centerline{30.12dB}
			\centerline{PatchDM\cite{ozdenizci2022restoring}}
		\end{minipage}
		\begin{minipage}[t]{0.1199\linewidth}			\centerline{\includegraphics[width=2.23cm]{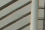}}
			\centerline{30.52dB}
			\centerline{Ours}
		\end{minipage}
		\begin{minipage}[t]{0.1199\linewidth}			\centerline{\includegraphics[width=2.23cm]{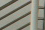}}
			\centerline{PSNR}       
			\centerline{GT}
		\end{minipage}
		\caption{Visual comparison on image raindrop removal. The PSNR values are computed on the whole images.}
		\label{fig_raindrop}
		\vspace{0.29cm}

		\centering
		\begin{minipage}[t]{0.1199\linewidth}			\centerline{\includegraphics[width=2.23cm]{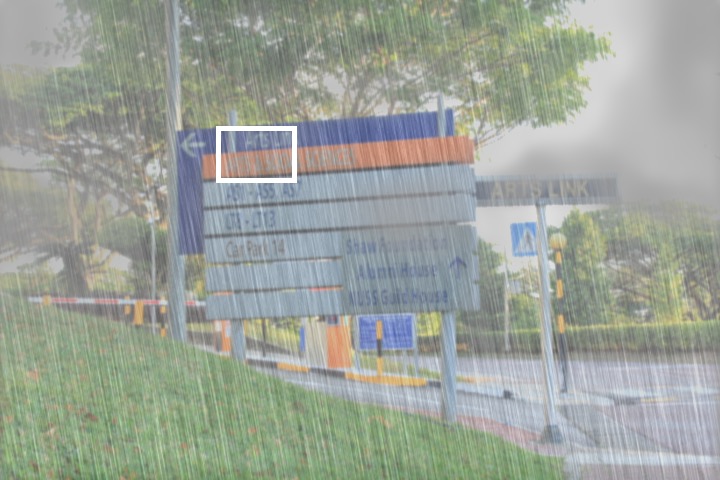}}
			\centerline{Degraded}
			\centerline{Image}
		\end{minipage}
		\begin{minipage}[t]{0.1199\linewidth}			\centerline{\includegraphics[width=2.23cm]{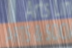}}
			\centerline{25.12dB}
			\centerline{Raindrop}
		\end{minipage}
		\begin{minipage}[t]{0.1199\linewidth}				\centerline{\includegraphics[width=2.23cm]{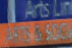}}
			\centerline{20.72dB}
			\centerline{HRGAN\cite{li2019heavy}}
		\end{minipage}
		\begin{minipage}[t]{0.1199\linewidth}			\centerline{\includegraphics[width=2.23cm]{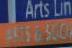}}
			\centerline{28.31dB}
			\centerline{MPRNet\cite{zamir2021multi}}
		\end{minipage}
		\begin{minipage}[t]{0.1199\linewidth}			\centerline{\includegraphics[width=2.23cm]{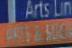}}
			\centerline{29.03dB}
			\centerline{TransWeather\cite{liu2019dual}}
		\end{minipage}
		\begin{minipage}[t]{0.1199\linewidth}			\centerline{\includegraphics[width=2.23cm]{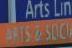}}
			\centerline{31.09dB}
			\centerline{PatchDM\cite{ozdenizci2022restoring}}
		\end{minipage}
		\begin{minipage}[t]{0.1199\linewidth}			\centerline{\includegraphics[width=2.23cm]{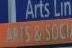}}
			\centerline{34.77dB}
			\centerline{Ours}
		\end{minipage}
		\begin{minipage}[t]{0.1199\linewidth}			\centerline{\includegraphics[width=2.23cm]{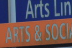}}
			\centerline{PSNR}       
			\centerline{GT}
		\end{minipage}
		\caption{Visual comparison on image rain steaks removal. The PSNR values are computed on the whole images.}
		\label{fig_outdoor}
		\vspace{0.29cm}
		
		\centering
		\begin{minipage}[t]{0.1199\linewidth}			\centerline{\includegraphics[width=2.23cm]{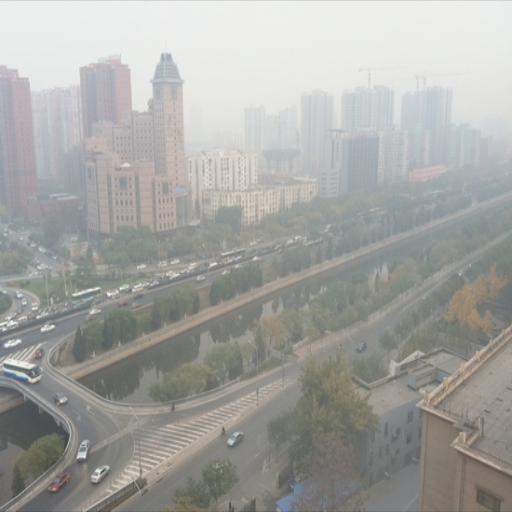}}
			\centerline{Degraded}
			\centerline{Image}
		\end{minipage}
		\begin{minipage}[t]{0.1199\linewidth}			\centerline{\includegraphics[width=2.23cm]{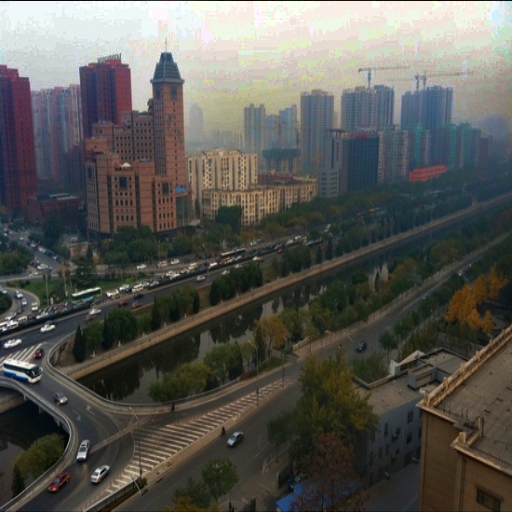}}
			\centerline{28.58dB}
			\centerline{DCP\cite{he2010single}}
		\end{minipage}
		\begin{minipage}[t]{0.1199\linewidth}			\centerline{\includegraphics[width=2.23cm]{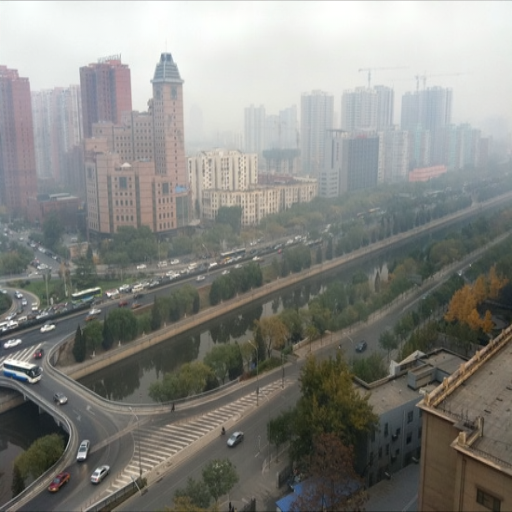}}
			\centerline{27.88dB}
			\centerline{GridDehazeNet\cite{liu2019griddehazenet}}
		\end{minipage}
		\begin{minipage}[t]{0.1199\linewidth}			\centerline{\includegraphics[width=2.23cm]{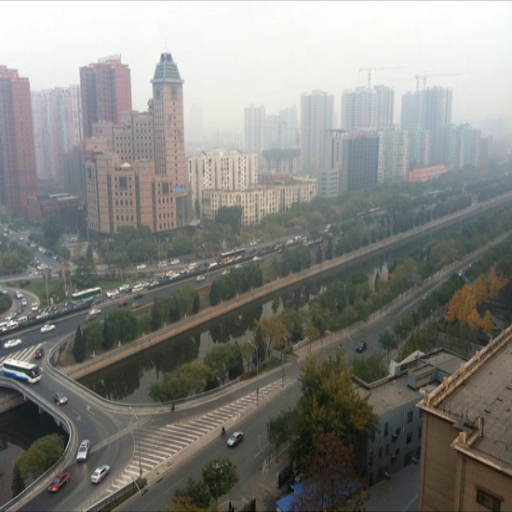}}
			\centerline{38.77dB}
			\centerline{MSBDN\cite{dong2020multi}}
		\end{minipage}
		\begin{minipage}[t]{0.1199\linewidth}			\centerline{\includegraphics[width=2.23cm]{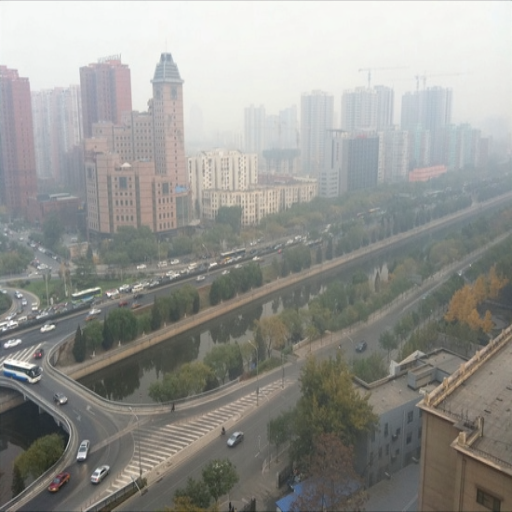}}
			\centerline{29.99dB}
			\centerline{FFA-Net\cite{qin2020ffa}}
		\end{minipage}
		\begin{minipage}[t]{0.1199\linewidth}			\centerline{\includegraphics[width=2.23cm]{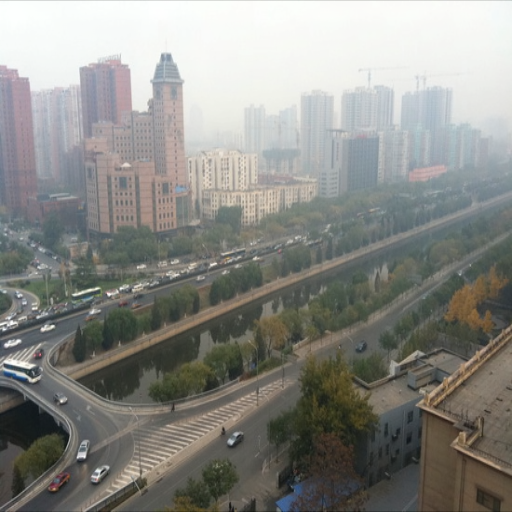}}
			\centerline{37.58dB}
			\centerline{DehazeFormer\cite{song2023vision}}
		\end{minipage}
		\begin{minipage}[t]{0.1199\linewidth}			\centerline{\includegraphics[width=2.23cm]{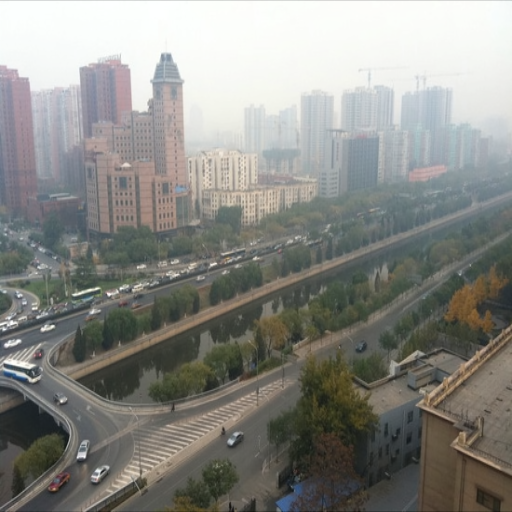}}
			\centerline{38.99dB}
			\centerline{Ours}
		\end{minipage}
		\begin{minipage}[t]{0.1199\linewidth}			\centerline{\includegraphics[width=2.23cm]{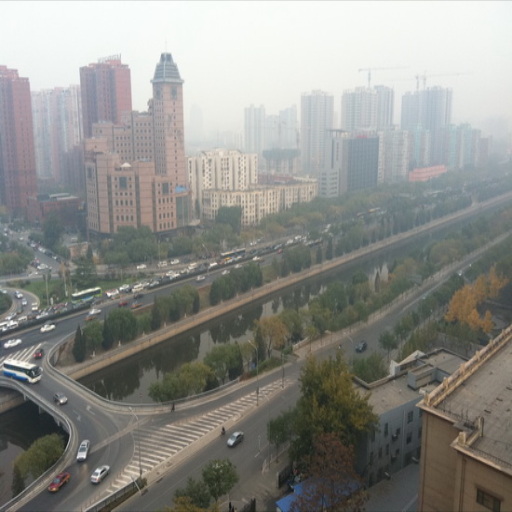}}
			\centerline{PSNR}       
			\centerline{GT}
		\end{minipage}
		\caption{Visual comparison on image dehazing.}
		\label{fig_dehaze}
		\vspace{0.29cm}

		\centering
		\begin{minipage}[t]{0.1199\linewidth}			
			\centerline{\includegraphics[width=2.23cm]{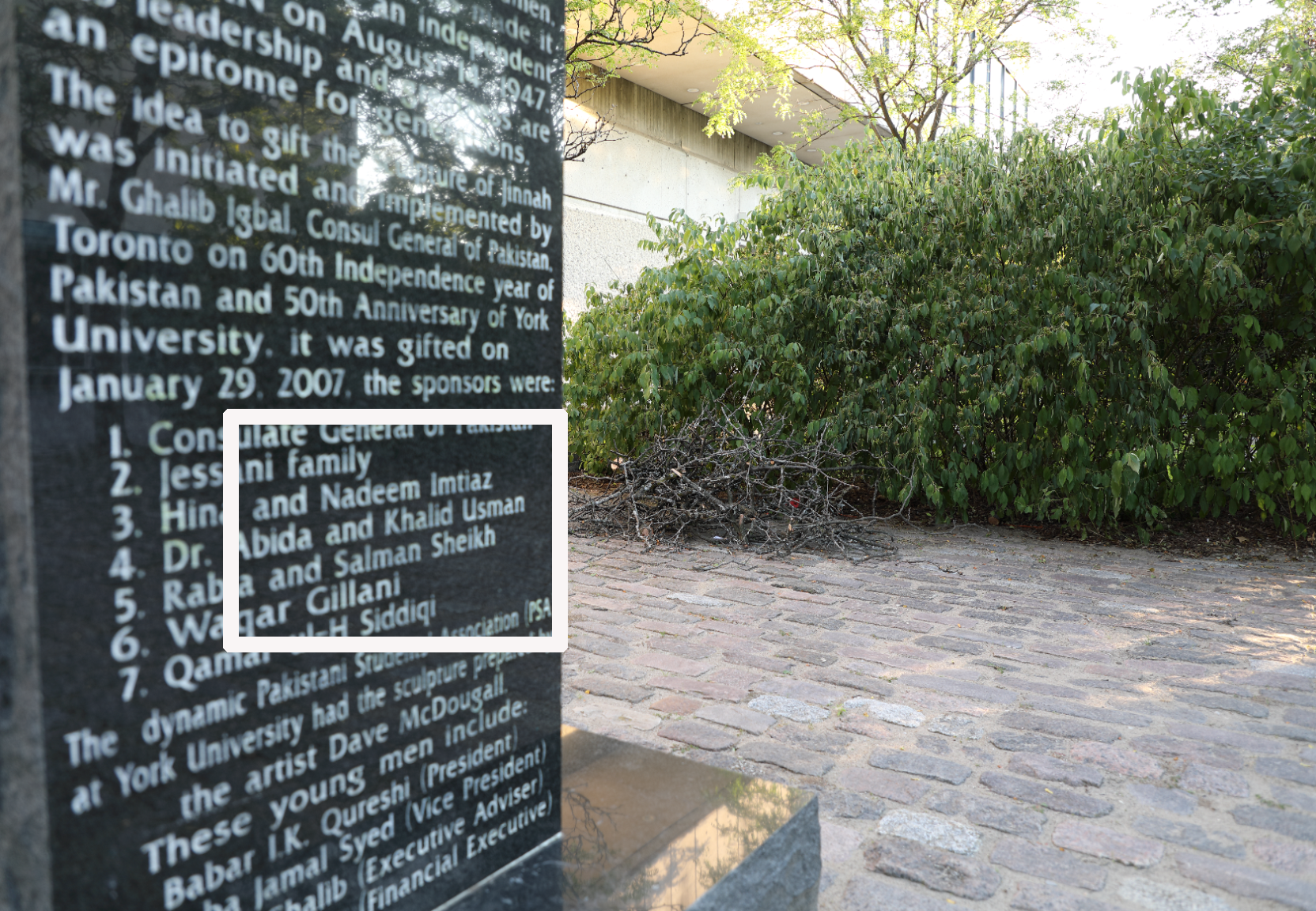}}
			\centerline{Degraded}
			\centerline{Image}
		\end{minipage}
		\begin{minipage}[t]{0.1199\linewidth}			
			\centerline{\includegraphics[width=2.23cm]{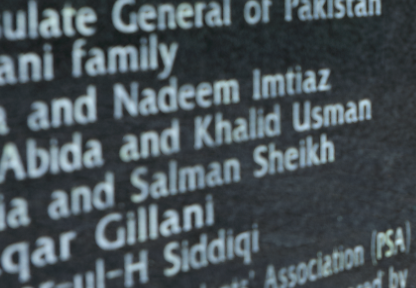}}
			\centerline{20.68dB}
			\centerline{Defocus}
		\end{minipage}
		\begin{minipage}[t]{0.1199\linewidth}			
			\centerline{\includegraphics[width=2.23cm]{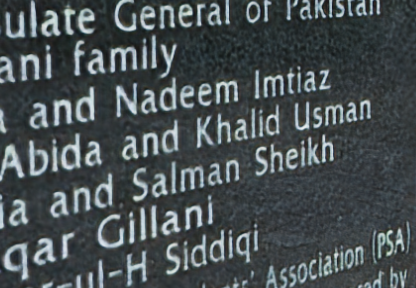}}
			\centerline{22.03dB}
			\centerline{DPDNet\cite{abuolaim2020defocus}}
		\end{minipage}
		\begin{minipage}[t]{0.1199\linewidth}			
			\centerline{\includegraphics[width=2.23cm]{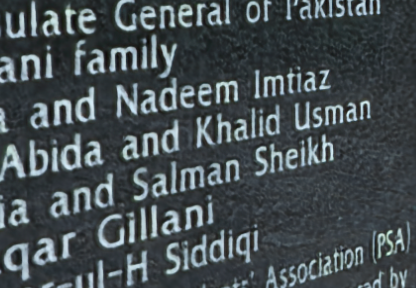}}
			\centerline{21.90dB}
			\centerline{IFAN\cite{lee2021iterative}}
		\end{minipage}
		\begin{minipage}[t]{0.1199\linewidth}			
			\centerline{\includegraphics[width=2.23cm]{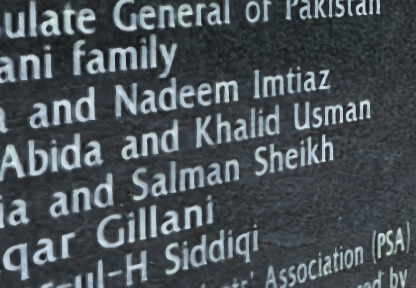}}
			\centerline{21.48dB}
			\centerline{Restormer\cite{zamir2022restormer}}
		\end{minipage}
		\begin{minipage}[t]{0.1199\linewidth}			
			\centerline{\includegraphics[width=2.23cm]{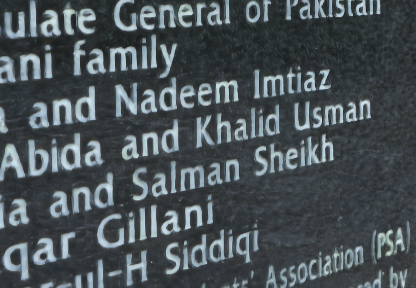}}
			\centerline{21.67dB}
			\centerline{PatchDM\cite{ozdenizci2022restoring}}
		\end{minipage}
		\begin{minipage}[t]{0.1199\linewidth}			
			\centerline{\includegraphics[width=2.23cm]{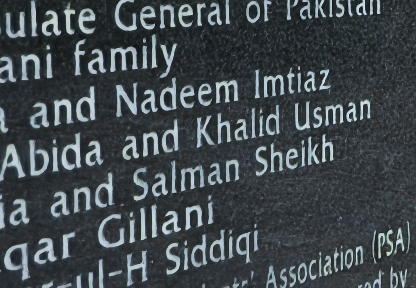}}
			\centerline{22.62dB}
			\centerline{Ours}
		\end{minipage}
		\begin{minipage}[t]{0.1199\linewidth}			
			\centerline{\includegraphics[width=2.23cm]{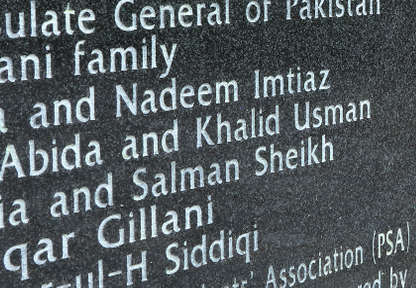}}
			\centerline{PSNR}
			\centerline{GT}
		\end{minipage}
		\caption{Visual comparison on image defocus deblurring. The PSNR values are computed on the whole images.}
		\label{fig_defocus}
		\vspace{0.29cm}

		\begin{minipage}[t]{0.1199\linewidth}			\centerline{\includegraphics[width=2.23cm]{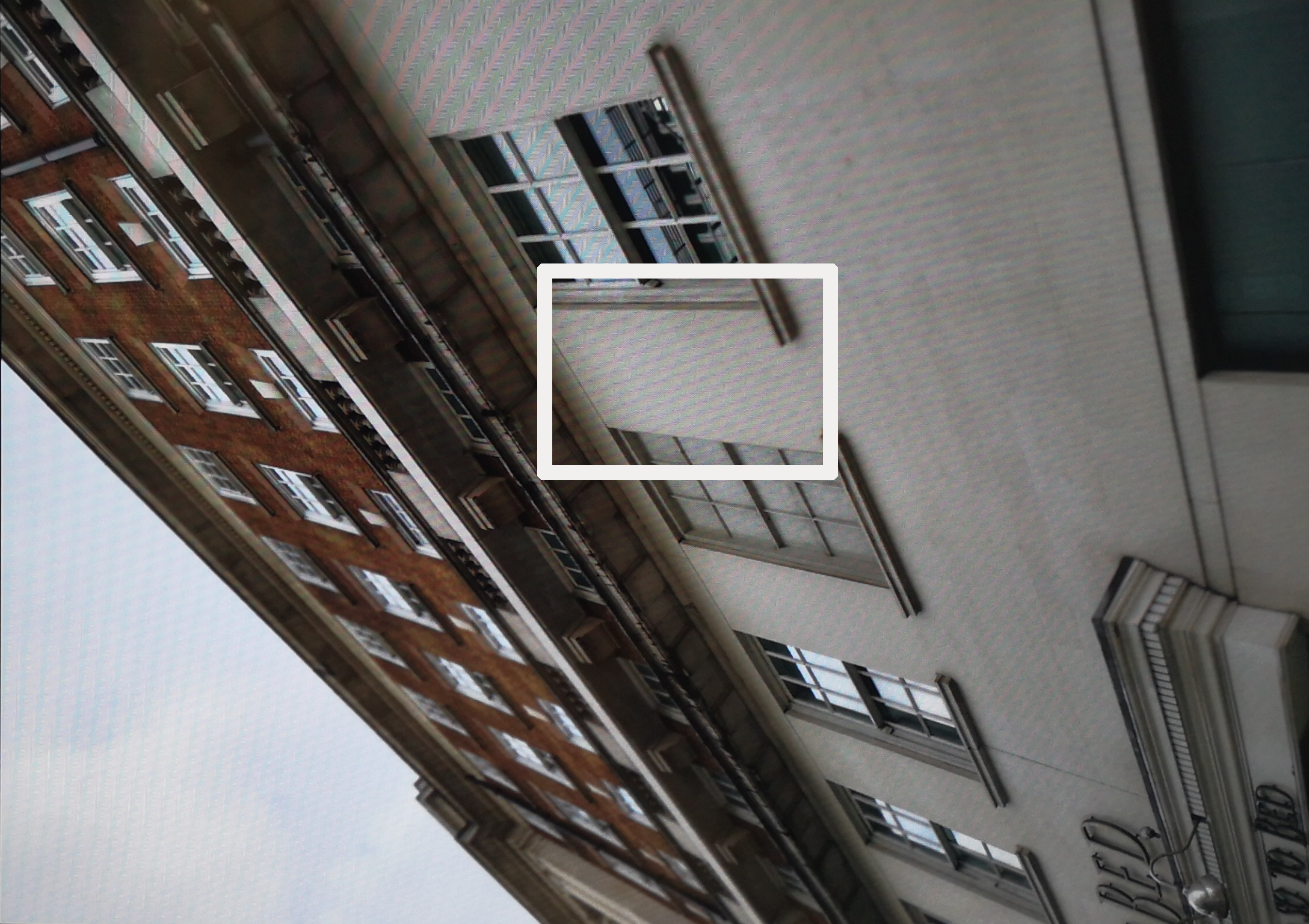}}
			\centerline{Degraded}
			\centerline{Image}
		\end{minipage}
		\begin{minipage}[t]{0.1199\linewidth}			\centerline{\includegraphics[width=2.23cm]{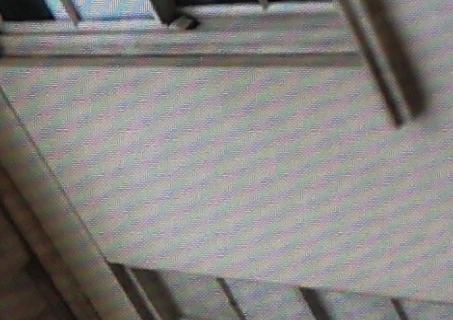}}
			\centerline{21.69dB}
			\centerline{Moiré}
		\end{minipage}
		\begin{minipage}[t]{0.1199\linewidth}			\centerline{\includegraphics[width=2.23cm]{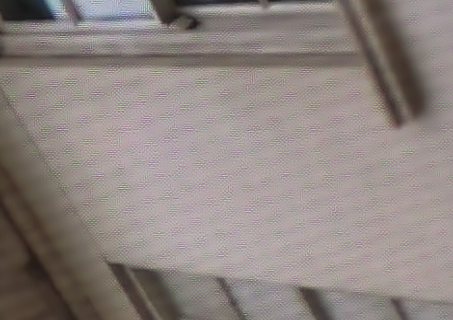}}
			\centerline{24.73dB}
			\centerline{FHDe2Net\cite{he2020fhde}}
		\end{minipage}
		\begin{minipage}[t]{0.1199\linewidth}			\centerline{\includegraphics[width=2.23cm]{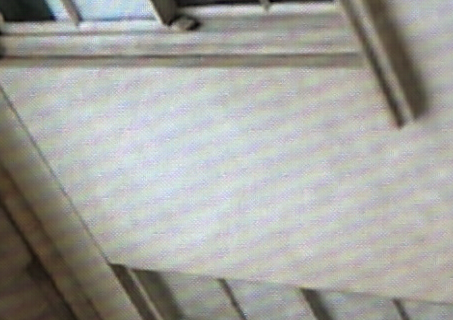}}
			\centerline{25.22dB}
			\centerline{WDNet\cite{liu2020wavelet}}
		\end{minipage}
		\begin{minipage}[t]{0.1199\linewidth}			\centerline{\includegraphics[width=2.23cm]{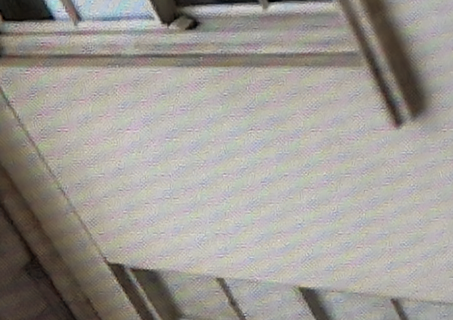}}
			\centerline{25.75dB}
			\centerline{ESDNet\cite{yu2022towards}}
		\end{minipage}
		\begin{minipage}[t]{0.1199\linewidth}			\centerline{\includegraphics[width=2.23cm]{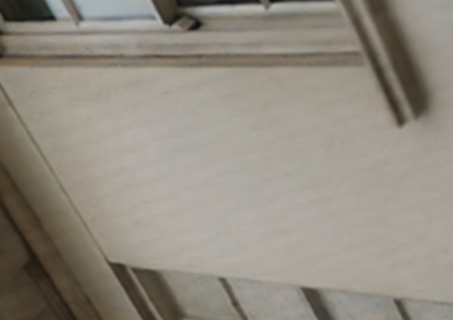}}
			\centerline{27.83dB}
			\centerline{PatchDM\cite{ozdenizci2022restoring}}
		\end{minipage}
		\begin{minipage}[t]{0.1199\linewidth}			\centerline{\includegraphics[width=2.23cm]{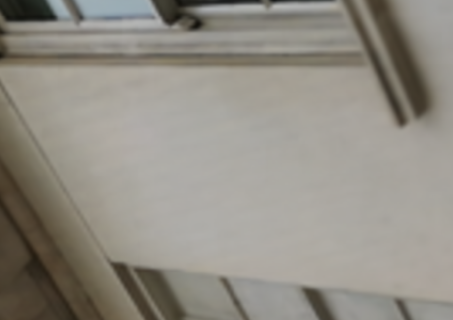}}
			\centerline{28.12dB}
			\centerline{Ours}
		\end{minipage}
		\begin{minipage}[t]{0.1199\linewidth}			\centerline{\includegraphics[width=2.23cm]{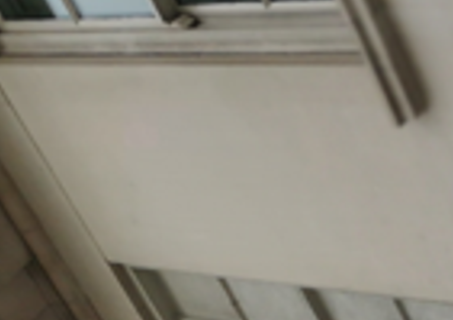}}
			\centerline{PSNR}
			\centerline{GT}
		\end{minipage}
		\caption{Visual comparison on image demoiréing. The PSNR values are computed on the whole images.}
		\label{fig_demoiré}
		\vspace{0.29cm}

		\begin{minipage}[t]{0.1199\linewidth}			\centerline{\includegraphics[width=2.22cm]{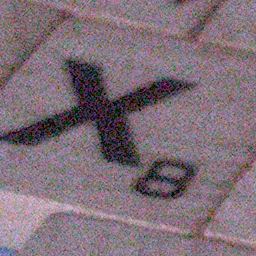}}
			\centerline{26.79dB}
			\centerline{Noisy Image}
		\end{minipage}
		\begin{minipage}[t]{0.1199\linewidth}			\centerline{\includegraphics[width=2.22cm]{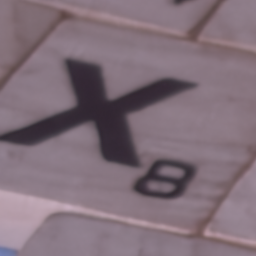}}
			\centerline{40.27dB}
			\centerline{MIRNet\cite{zamir2020learning}}
		\end{minipage}
		\begin{minipage}[t]{0.1199\linewidth}			\centerline{\includegraphics[width=2.22cm]{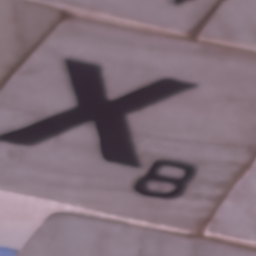}}
			\centerline{40.42dB}
			\centerline{MPRNet\cite{zamir2021multi}}
		\end{minipage}
		\begin{minipage}[t]{0.1199\linewidth}			\centerline{\includegraphics[width=2.22cm]{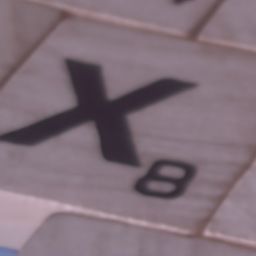}}
			\centerline{40.64dB}
			\centerline{Uformer\cite{wang2022uformer}}
		\end{minipage}
		\begin{minipage}[t]{0.1199\linewidth}			\centerline{\includegraphics[width=2.22cm]{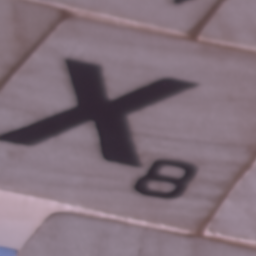}}
			\centerline{40.67dB}
			\centerline{Restormer\cite{zamir2022restormer}}
		\end{minipage}
		\begin{minipage}[t]{0.1199\linewidth}			\centerline{\includegraphics[width=2.22cm]{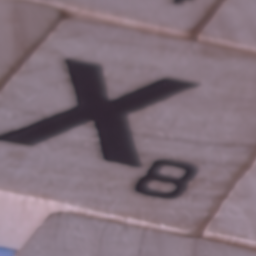}}
			\centerline{41.21dB}
			\centerline{PatchDM\cite{ozdenizci2022restoring}}
		\end{minipage}
		\begin{minipage}[t]{0.1199\linewidth}			\centerline{\includegraphics[width=2.22cm]{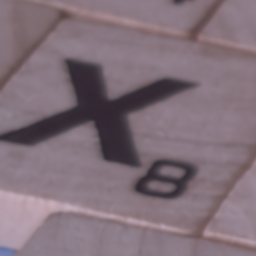}}
			\centerline{42.47dB}
			\centerline{Ours}
		\end{minipage}
		\begin{minipage}[t]{0.1199\linewidth}			\centerline{\includegraphics[width=2.22cm]{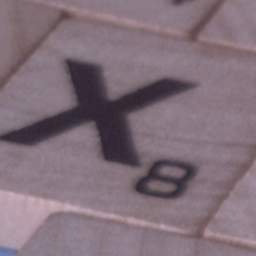}}
			\centerline{PSNR}
			\centerline{GT}
		\end{minipage}
		\caption{Visual comparison on real image denoising.}
		\label{fig_denoise}
	\end{figure*}
	
	\subsubsection{Weight of Loss Functions}

	In the process of optimizing our model, the weighting parameter \(\lambda\) is introduced in the loss function defined in Eq.~\ref{total_loss}. This parameter serves as a trade-off between the losses associated with the low-frequency wavelet bands, denoted as \(L_{\text{simple}}\), and the high-frequency bands, represented as \(L_1\). Fig.~\ref{fig_lambda} visually presents the experimental results, portraying the performance of our model across different iterations for various choices of \(\lambda\). This evaluation is conducted on the London's Buildings dataset \cite{liu2020wavelet} for image demoiréing. 
	From the results, we can obtain the following observations. First, the choice of $\lambda$ impacts the model's convergence rate. A smaller $\lambda$ (e.g., 0.01) leans towards slower convergence due to the focus on low-frequency bands. Conversely, higher values of $\lambda$ like 1 or 10 accelerate the convergence. Second, despite the variance in convergence, the final restoration quality, measured in PSNR, remains almost consistent across a large range of $\lambda$ values (e.g., [0.1, 10]), indicating that our model is not very sensitive to $\lambda$. However, too small or too large $\lambda$ would harm the performance, which reveals the intricate relationship between wavelet spectrum learning and the balance of high- and low- frequency representations. An inappropriate weighting might lead the model to overly focus on either the detailed high frequencies or the coarse low frequencies. In all the following experiments, $\lambda$ is set to 1.

	\subsubsection{Conditional Sampling Variability}

	To delve deeper into WaveDM's sampling capability from the conditional distribution, we conduct an experiment where we vary the seed to produce different samples while maintaining a constant degraded image as the condition.
	Fig.~\ref{fig_seed} exhibits two groups of generated images for six distinct seeds. To offer a quantitative measure, we also compute the PSNR between each of these samples and the ground truth of the degraded image.
	The visual inspection of these images along with the quantitive comparison brings an observation that the differences between the samples, even though generated using different seeds, are extremely subtle both visually and quantitively. On the contrary, with different conditions (i.e., different degraded images), the results are generated differently and guided by the conditions.
	This clearly shows WaveDM's consistent ability to produce high-quality restorations, regardless of the minor variabilities introduced by different seeds.
	
	\subsection{Comparison with State-of-the-Art Methods}
	We evaluate our WaveDM with other state-of-the-art (SOTA) methods on twelve benchmark datasets.

	All results are obtained either by copying from their papers or retraining and testing with their official code and released pretrained models. We re-implement PatchDM for this task as the baseline. The best and second best values are indicated in \textbf{bold} and \underline{underlined}, respectively. Since the default PatchDM cuts images into multiple patches of size $64\times64$, denoted as PatchDM$_{64}$, its memory used keeps unchanged across different image sizes. PatchDM also provides another version PatchDM$_{128}$, which cuts images into $128\times128$ patches. Besides, as our noise estimation network keeps the same as the baseline PatchDM for a fair comparison, the extra parameters only come from HFRM. The FID, processing time, parameter count, and the memory usage of method All-in-One \cite{li2020all} cannot be obtained due to unavailability of its code.

	\begin{table}[h]
		\small
		\renewcommand\arraystretch{1.1}
		\begin{center}
			\caption{Quantitative comparison with SOTA methods for Gaussian grayscale image denoising on three common benchmarks.}\label{table_gray_denoising}
			\resizebox{0.98\linewidth}{!}{
				\begin{tabular}{c|ccc|ccc|ccc}
					\toprule
					\multirow{2}*{Methods}&\multicolumn{3}{c|}{Set12\cite{zhang2017beyond}}&\multicolumn{3}{c|}{BSD68\cite{martin2001database}}&\multicolumn{3}{c}{Urban100\cite{huang2015single}}\\
					\cmidrule{2-10}
					~&$\sigma=15$&$\sigma=25$&$\sigma=50$&$\sigma=15$&$\sigma=25$&$\sigma=50$&$\sigma=15$&$\sigma=25$&$\sigma=50$\\
					\toprule
					MWCNN\cite{liu2018multi}& 33.15&30.79&27.74&31.86&29.41&26.53&33.17&30.66&27.42\\
					DeamNet\cite{ren2021adaptive}&33.19&30.81&27.74&31.91&29.44&26.54&33.37&30.85&27.53\\
					DAGL\cite{mou2021dynamic}&33.28&30.93&27.81&31.93&29.46&26.51&\underline{33.79}&31.39&27.97\\
					SwinIR\cite{liang2021swinir}&33.36&31.01&27.91&\textbf{31.97}&29.50&26.58&33.70&31.30&27.98\\
					Restormer\cite{zamir2022restormer}&\underline{33.42}&\underline{31.08}&\underline{28.00}&\underline{31.96}&\underline{29.52}&\textbf{26.62}&\underline{33.79}&\underline{31.46}&\textbf{28.29}\\
					
					Ours&\textbf{33.75}&\textbf{31.47}&\textbf{28.44}&31.95&\textbf{29.58}&\underline{26.60}&\textbf{33.92}&\textbf{31.86}&\underline{28.21}\\
					
					\bottomrule
				\end{tabular}
			}
			
		\end{center}
	\end{table}
	
	\begin{table}[ht]
		\small
		\renewcommand\arraystretch{1.1}
		\begin{center}
			\caption{Quantitative comparison with SOTA methods for Gaussian color image denoising on three common benchmarks.}\label{table_color_denoising}
			\resizebox{0.98\linewidth}{!}{
				\begin{tabular}{c|ccc|ccc|ccc}
					\toprule
					\multirow{2}*{Methods}&\multicolumn{3}{c|}{CBSD68\cite{martin2001database}}&\multicolumn{3}{c|}{Kodak24\cite{zhang2011color}}&\multicolumn{3}{c}{Urban100\cite{huang2015single}}\\
					\cmidrule{2-10}
					~&$\sigma=15$&$\sigma=25$&$\sigma=50$&$\sigma=15$&$\sigma=25$&$\sigma=50$&$\sigma=15$&$\sigma=25$&$\sigma=50$\\
					\toprule
					IRCNN\cite{zhang2017learning}&33.86&31.16&27.86&34.69&32.18&28.93&33.78&31.20&27.70\\
					DnCNN\cite{zhang2017beyond}&33.90&31.24&27.95&34.60&32.14&28.95&32.98&30.81&27.59\\
					FFDNet\cite{zhang2018ffdnet}&33.87&31.21&27.96&34.63&32.13&28.98&33.83&31.40&28.05\\
					DRUNet\cite{zhang2021plug}&34.30&31.69&28.51&35.31&32.89&29.86&34.81&32.60&29.61\\
					Restormer\cite{zamir2022restormer}&\underline{34.39}&\underline{31.78}&\underline{28.59}&\textbf{35.44}&\textbf{33.02}&\underline{30.00}&\underline{35.06}&\textbf{32.91}&\underline{30.02}\\
					Ours&\textbf{34.85}&\textbf{31.81}&\textbf{28.78}&\underline{35.41}&\underline{32.97}&\textbf{30.23}&\textbf{35.31}&\underline{32.89}&\textbf{30.22}\\
					\bottomrule
				\end{tabular}
			}
			
		\end{center}
	\end{table}
	
	\begin{table}[t]
		\scriptsize
		\renewcommand\arraystretch{1.1}
		\begin{center}
			\caption{Comparison with latent diffusion implementation.}\label{table_ldm}
			\resizebox{0.9\linewidth}{!}{
				\begin{tabular}{c|ccc|cc}
					\toprule
					\multirow{2}*{Implementation}&\multicolumn{3}{c|}{Testing}&\multicolumn{2}{c}{Upper Bound}\\
					\cmidrule{2-6}
					~&PSNR$\uparrow$&SSIM$\uparrow$&Time$\downarrow$&PSNR$\uparrow$&SSIM$\uparrow$\\
					\toprule
					WaveDM& 31.39dB&0.943&0.81s&56.38dB&0.999\\
					LDM&24.31dB&0.873&1.05s&25.82dB&0.895\\
					\bottomrule
				\end{tabular}
			}
			
		\end{center}
	\end{table}

	\subsubsection{Image Raindrop Removal}
	In our evaluation on the RainDrop dataset\cite{qian2018attentive}, various methods are analyzed for their raindrop removal efficiency, with the results tabulated in Table~\ref{table_sota}. While PatchDM\cite{ozdenizci2022restoring} slightly edges out in terms of PSNR (a marginal 0.06 dB advantage over WaveDM), its practicality is limited due to the extensive inference time involved in patch processing. WaveDM, in contrast, demonstrates comparable performance in a fraction of the time, proving its efficiency in dealing with challenging conditions like heavy raindrop obstruction, as substantiated in Fig.~\ref{fig_raindrop}.
	
	\subsubsection{Image Rain Streaks Removal}
	Rain streaks present a different challenge compared to raindrops. 
	Fig.~\ref{fig_outdoor} exhibits the visual results on the Outdoor-rain dataset \cite{li2019heavy}, which validate that WaveDM shows an excellent capability in removing heavy rain streaks without compromising image details. Besides, quantitative evaluations are presented in Table~\ref{table_sota}, also demonstrating WaveDM's better performance against others' with a competitive processing time.
	
	\subsubsection{Image Dehazing}
	In addition to the rainy scenarios, we also apply our method to another adverse weather condition, haze. We select 6000 images from the training set SOTS \cite{li2018benchmarking} that contains	over 70000 images for training, and evaluate our model on the  SOTS-Outdoor benchmark. The quantitive results shown in Table~\ref{table_sota} demonstrate that WaveDM achieves the best PSNR and SSIM. Besides, WaveDM obtains 2.8 in terms of FID, showing high fidelity of restored samples. The visual samples presented in Fig.~\ref{fig_dehaze} evidence WaveDM's dehazing performance, including the image's clarity, sharpness, and color preservation.
	
	\subsubsection{Image Defocus Deblurring}
	Our experiments on the DPDD dataset\cite{abuolaim2020defocus} for both single and dual-pixel defocus deblurring, as captured in Table~\ref{table_sota} and Fig.~\ref{fig_defocus}, reveal WaveDM's superior performance over other SOTA methodologies with a comparable inference time to the one-pass methods.
	
	\subsubsection{Image Demoiréing}
	In dealing with moiré patterns in images, especially those from the London's Buildings dataset \cite{liu2020wavelet}, WaveDM proves to be a strong competitor. While diffusion models generally perform better than one-pass methods, WaveDM distinguishes itself by obtaining superb performance with quick inference, matching the speed of one-pass systems, as evidant in Table~\ref{table_sota} and Fig.~\ref{fig_demoiré}.

	\subsubsection{Real Image Denoising}
	On the SIDD dataset\cite{abdelhamed2018high} for real image denoising tasks, WaveDM's effectiveness is further confirmed. It not only outperforms other one-pass methods but also matches the performance of diffusion-based methods such as PatchDM. The images in Fig.~\ref{fig_denoise} and numerical evaluations in Table~\ref{table_sota} clearly showcase its strengths.

	\subsubsection{Gaussian Image Denoising}
	In addition to real image denoising, we also apply WaveDM to Gaussian image denoising. For grayscale image denoising, we employ three widely-used benchmark datasets: Set12\cite{zhang2017beyond}, BSD68\cite{martin2001database}, and Urban100\cite{huang2015single}. The quantitative results, presented in Table \ref{table_gray_denoising}, show that our method overall outperforms other competing methods across different noise levels. Specifically, for noise level $\sigma=50$, our method achieves a remarkable PSNR of 28.44dB on Set12, which is notably higher than other methods.
	Similarly, for color image denoising, our experiments span across datasets CBSD68\cite{martin2001database}, Kodak24\cite{zhang2011color}, and Urban100\cite{huang2015single}. The results, detailed in Table \ref{table_color_denoising}, further solidify our method's superior performance. For instance, on the Urban100 dataset at noise level $\sigma=50$, our model achieves an impressive PSNR of 30.22dB, surpassing all competitors.
	These experimental results offer solid evidence that WaveDM is not only robust to varying degrees of Gaussian noise but also consistently outperforms SOTA methods, emphasizing its effectiveness and adaptability.

	\subsection{Comparison with Latent Diffusion Implementation}

	To further demonstrate the efficiency and effectiveness of WaveDM, which employs wavelet transform for image size reduction and diffusion modeling in the wavelet domain, we conduct a comparative experiment on the Outdoor-rain dataset\cite{li2019heavy} with the Latent Diffusion Model (LDM) \cite{rombach2022high} implementation, a method that can also reduce image size using VAE-based subsampling. Specifically, the images, sized $720\times480\times3$, are processed by a 4-downsampled pretrained VAE from \cite{rombach2022high} to be transformed into a latent space. These transformed images are then used as input for the latent diffusion model (LDM), which keeps its architecture the same as WaveDM's. The results of this comparative evaluation are presented in Table \ref{table_ldm}, in which ``Upper Bound" gives the maximum results WaveDM and LDM can achieve, where the PSNR and SSIM values are computed by directly applying either the wavelet transform (and its inverse) or the VAE’s encoder-decoder on clean images (ground truth), without any diffusion processing. From the table, we observe that the VAE's reconstruction restricts LDM's restoration capability. Additionally, LDM tends to be slightly slower than WaveDM, since the wavelet transformation is inherently more efficient than VAE processing.
	In conclusion, WaveDM demonstrates superior restoration and efficiency compared to the LDM alternative.

	\section{Conclusion and Limitation}
	This paper proposes a wavelet-based diffusion model (WaveDM) to reduce the inference time of diffusion-based models for image restoration. WaveDM learns the distribution in the wavelet domain of clean images, which saves a lot of time in each step of sampling. In addition, an efficient conditional sampling technique is developed from experiments to reduce the total sampling steps to around 5.
	Experiments on twelve image datasets validate that our WaveDM achieves SOTA performance with the efficiency that is over $100\times$ faster than the previous diffusion-based SOTA PatchDM and is also comparable to traditional one-pass methods.
	
	The major limitation is that WaveDM requires millions of training iterations for several days, especially for large-scale datasets, which is left to deal with in future work.

	{\small
		\bibliographystyle{ieee_fullname}
		\bibliography{egbib}
	}

	\newpage

	\vfill
	
\end{document}